%% file: counterfactual-llms.tex
\documentclass{article}
\pdfoutput=1

\usepackage[utf8]{inputenc} 
\usepackage[T1]{fontenc}    
\PassOptionsToPackage{hyphens}{url} 
\usepackage[hyperfootnotes=false]{hyperref}
\usepackage{booktabs}       
\usepackage{amsfonts}       
\usepackage{nicefrac}       
\usepackage{microtype}      
\usepackage{xcolor}         
\usepackage[noend,ruled]{algorithm2e}

\usepackage{graphicx}
\usepackage{enumerate}
\usepackage{caption}
\usepackage{subcaption}
\usepackage{comment}
\usepackage[export]{adjustbox}
\usepackage{mathtools}

\usepackage{authblk}
\usepackage{fullpage}

\usepackage{enumitem}

\usepackage[sort&compress, numbers]{natbib}

\usepackage{tikz}
\usepackage{xcolor}
\usepackage{tcolorbox}

\usepackage{Definitions}
\usepackage{dsfont}

\newlength{\inlineheight}
\newcommand{\UnnumberedFootnote}[1]{{\def\thefootnote{}\footnote{#1}
\addtocounter{footnote}{-1}}}

\title{Counterfactual Token Generation in Large Language Models}

\author{Ivi Chatzi$^{*}$}
\author{Nina Corvelo Benz$^{*}$}
\author{Eleni Straitouri$^{*}$}
\author{Stratis Tsirtsis$^{*}$}
\author{Manuel~Gomez-Rodriguez}

\affil{Max Planck Institute for Software Systems \\ Kaiserslautern, Germany \\
\{ichatzi, ninacobe, estraitouri, stsirtsis, manuel\}@mpi-sws.org}

\date{}

\begin{document}

\maketitle

\UnnumberedFootnote{$^{*}$Authors contributed equally and are listed in alphabetical order.}

\begin{abstract}
\input{000abstract.tex}
\end{abstract}

\section{Introduction}
\label{sec:intro}
\input{010introduction.tex}

\section{A Causal Model of Token Generation}
\label{sec:scm}
\input{020scm.tex}

\section{Counterfactual Token Generation Using Gumbel-Max SCMs}
\label{sec:counterfactual}
\input{030implementation.tex}

\section{Experiments}
\label{sec:experiments}
\input{040experiments.tex}

\section{Discussion and Limitations}
\label{sec:discussion}
\input{050discussion.tex}

\section{Conclusions}
\label{sec:conclusions}
\input{060conclusions.tex}

\vspace{2mm}
\xhdr{Acknowledgements} 
Gomez-Rodriguez acknowledges support from the European Research Council (ERC) under the European Union'{}s Horizon 2020 research and innovation programme (grant agreement No. 945719).

{ 
\small
\bibliographystyle{unsrt}
\bibliography{counterfactual-llms}
}

\clearpage
\newpage

\appendix

\input{070appendix.tex}

\end{document}

%% file: 000abstract.tex
\textit{``Sure, I am happy to generate a story for you:
Captain Lyra stood at the helm of her trusty ship, the Maelstrom's Fury, gazing out at the endless sea. [...] Lyra's eyes welled up with tears as she realized the bitter truth -- she had sacrificed everything for fleeting riches, and lost the love of her crew, her family, and herself.''}
Although this story, generated by a large language model, is captivating, one may wonder---how would the story have unfolded if the model had chosen ``Captain Maeve'' as the protagonist instead?
We cannot know. State-of-the-art large language models are stateless---they maintain no internal memory or state. 
Given a prompt, they generate a sequence of tokens as an output using an autoregressive process. 
%
As a consequence, they cannot reason about counterfactual alternatives to tokens they have generated in the past. 
%
In this work, our goal is to enhance them with this functionality.
%
To this end, we develop a causal model of token generation that builds upon the Gumbel-Max structu\-ral causal model.
%
Our model allows any large language model to perform counterfactual token generation at almost no cost in comparison with vanilla token generation, it is embarrassingly simple to implement, and it does not require any fine-tuning nor prompt engineering.
%
We implement our model on \texttt{Llama 3 8B-Instruct} and \texttt{Ministral-8B-Instruct}, and conduct a qualitative and a quantitative analysis of counterfactually generated text. We conclude with a demonstrative application of counterfactual token generation for bias detection, unveiling interesting insights about the model of the world constructed by large language models.

%% file: 010introduction.tex
Reasoning about ``what might have been'', about alternatives to our own past actions, is a landmark of human~intelli\-gence~\citep{roese1997counterfactual,byrne2007precis,van2015cognitive}.
This type of reasoning, known as counterfactual reasoning, has been shown to play a significant role in the ability that humans have to learn from limited past experience and improve their decision making skills over time~\citep{epstude2008functional,markman2008counterfactual,roese2017functional},
it provides the basis for creativity and insight~\citep{sternberg1989if},
and it is tightly connected to the way we attribute causality and responsibility~\citep{lagnado2013causal,gerstenberg2021counterfactual,xiang2023actual,tsirtsis2024towards}.
Can currently available large language models (LLMs) conduct counterfactual reasoning about alternatives to their own outputs?
In this work, we argue that they cannot, by design.

Currently available LLMs are stateless---they maintain no internal memory or state.
Given an input prompt, they generate a sequence of tokens\footnote{Tokens are the units that make up sentences and paragraphs. Examples of tokens include (sub-)words, symbols, numbers, and special end-of-sequence tokens.} as output using an autoregressive process~\cite{bengio2000neural,radford2019language}. 
At each time step, they first use a neural network to map the prompt and the (partial) sequence of tokens generated so far to a token distribution. 
Then, they use a sampler to draw the next token at random from the token distribution.\footnote{Multiple lines of evidence suggest that, if an LLM is forced to output tokens deterministically, its performance worsens~\citep{holtzman2019curious}.}
Finally, they append the next token to the (partial) sequence of tokens, and continue until a special end-of-sequence token is sampled.
To understand why this autoregressive process is insufficient to
reason counterfactually about alternatives to a previously generated sequence of tokens, we will use an illustrative example.

\begin{figure}[t]
    \captionsetup[subfigure]{justification=centering}
    \centering
    \subcaptionbox{Original generation~\label{fig:example-f}}{
        \includegraphics[width=0.3\textwidth]{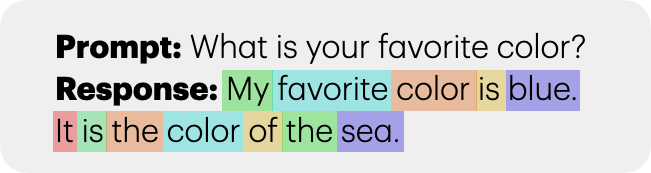}
    }
    \hspace{10mm}
    \subcaptionbox{Interventional generation\\with unmodified input~\label{fig:example-p}}{
        \includegraphics[width=0.3\textwidth]{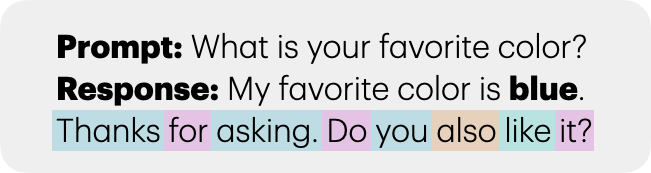}
    }
    \subcaptionbox{Interventional generation\\with modified input~\label{fig:example-int}}{
        \includegraphics[width=0.3\textwidth]{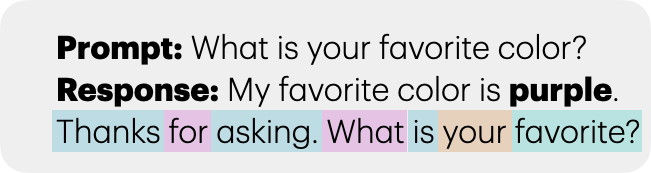}
    }
    \hspace{10mm}
    \subcaptionbox{Counterfactual generation\\with modified input~\label{fig:example-cf}}{
        \includegraphics[width=0.3\textwidth]{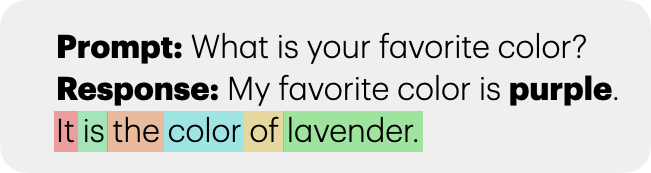}
    }
    \caption{\textbf{Illustrative examples of autoregressive token generation.} 
    In all panels, plain text indicates the input provided to the LLM and highlighted text indicates the output generated by the model.
    Each token in the output sequence is highlighted in a different color to represent the (stochastic) state of the sampler.\protect\footnotemark\, 
    %
    Panel (a) shows an LLM's output to a user's prompt using vanilla autoregressive token generation. 
    Panels (b, c) show an LLM's output to an input comprising a user's prompt and an unmodified/modified part of the original output from Panel (a) using vanilla autoregressive token generation. 
    Panel (d) shows an LLM's counterfactual output to an input comprising a user's prompt and a modified part of the output from Panel (a) using autoregressive token generation augmented with the Gumbel-Max SCM. 
    }\label{fig:example}
\end{figure}

\footnotetext{In Section~\ref{sec:scm}, we formally define the state of the sampler as an exogenous noise variable in an SCM.}

Consider that we ask an LLM to share its favorite color, as shown in Figure~\ref{fig:example-f}.
Had the LLM chosen a different color (\eg, purple instead of blue), what would the rest of its output have been?
To answer such a counterfactual question, we need to implement two actions: 
(i) modify the (partial) sequence of tokens fed to the neural network used by the LLM and (ii) compel the sampler used by the LLM to \emph{behave exactly as it did} in the original generation.
Using currently available LLMs, we can readily implement the first action, which can be viewed as a causal intervention~\cite{pearl2009causality, peters2017elements}. We just need to replace ``blue'' with ``purple'' in the (partial) sequence of tokens fed to the neural network.
However, we cannot easily implement the second action, because the sampler does not specify how it \emph{would have behaved} after taking the first action \emph{while keeping everything else equal}.
In fact, note that, if we provide the (modified) partial sequence up to and including the world ``blue'' (``purple'') as input to the LLM, there is no way to ensure that the LLM will generate an output that matches (the structure of) the original output, 
as shown in Figures~\ref{fig:example-f},~\ref{fig:example-p} and~\ref{fig:example-int}.\footnote{Note that using the same random seed is not sufficient because the number of tokens of the input in Figure~\ref{fig:example-f} and the number of tokens of the inputs in Figures~\ref{fig:example-p} and~\ref{fig:example-int} differ.}

\xhdr{Our contributions}
%
Our key idea is to augment the autoregressive process of token generation underpinning an LLM, particularly the sampler used in the process, using a structural causal model (SCM)~\cite{pearl2009causality}. More specifically, we define the sampler through a causal mechanism that receives as input the distribution of the next token and a set of noise values, which determine the sampler's (stochastic) state.
Importantly, the use of a causal mechanism specifies how the sampler would have behaved under an intervention on the distribution of the next token and thus allows us to answer counterfactual questions about a previously generated sequence of tokens, as shown in Figure~\ref{fig:example-cf}.
Further, to instantiate our model, we use the Gumbel-Max SCM~\cite{oberst2019counterfactual}, an SCM shown to satisfy a desirable counterfactual stability property which, in the context of token generation, favors counterfactual output sequences that share similarities with the original sequence. 
Along the way, we also introduce an efficient implementation of the augmented autoregressive process that can generate counterfactual tokens at almost no cost in comparison with vanilla token generation.
%
As a proof of concept, we implement our model on \texttt{Llama 3 8B-Instruct} and \texttt{Ministral-8B-Instruct}, and we conduct experiments to qualitatively and quantitatively analyze the similarity between an LLM's original output and the one generated via counterfactual token generation. Additionally, we demonstrate the use of our methodology for bias detection, unveiling interesting insights about the model of the world constructed by large language models. 
We conclude with a comprehensive discussion of the limitations of our model, including additional avenues for applications.
%
An open-source implementation of our model on \texttt{Llama 3 8B-Instruct} and \texttt{Ministral-8B-Instruct} is available at \url{https://github.com/Networks-Learning/counterfactual-llms}.

\xhdr{Further related work} 
%
Our work is most closely related to a line of work on counterfactual text gen\-er\-a\-tion\mbox{~\citep{qin2019counterfactual,qin2020back,hao2021sketch,chen2022unsupervised,wang2024survey,wang2024beyond,nguyen2024llms,zellers2019hellaswag,li2023prompting,nguyen2024ceval,gat2024faithful}}. 
In this line of work, given pairs of factual statements and interventions over these statements, the goal is to generate counterfactual statements that match those made by humans---counterfactual statements that are consistent with the underlying model of the world shared by humans.
To this end, existing methods typically fine-tune an LLM using a dataset comprising factual statements, interventions over these statements, and counterfactual statements made by humans.
In contrast, in our work, our goal is to generate~coun\-ter\-factual statements that are consistent with the underlying model of the world constructed by a given LLM~\citep{li2021implicit, li2022emergent, patel2022mapping,vafa2024evaluating}.
%
In this context, our work also relates to a rapidly increasing number of empirical studies assessing the ability of LLMs to answer questions that require counterfactual reasoning~\citep{frohberg2022crass,jin2023cladder,kiciman2023causal,pawlowski2023answering,jiang2023large,betti2023relevance,liu2023magic,miao2023generating,wu2023reasoning,nie2024moca,ortu2024competition,liu2024large}.
Here, the LLMs are typically evaluated using multiple choice questions about a given set of factual and counterfactual statements.
However, similarly as in the line of work on counterfactual text generation discussed previously, the counterfactual statements are made by humans.

The works by Ravfogel et al.~\cite{ravfogel2025true} and Bynum and Cho~\cite{bynum2024language}, which have been undertaken concurrently and independently of our work, also use SCMs to model the autoregressive process underpinning LLMs as we do.
The work by Ravfogel et al.~\cite{ravfogel2025true} is most closely related to ours; they model autoregressive generation using the Gumbel-Max SCM and generate counterfactual strings to visualize and analyze the effects of interventions within (the network of) an LLM.
However, in contrast to our work, they do not explicitly implement the sampler of an LLM as an SCM and use the same set of noise values during factual and counterfactual generation, but they sample the noise values using a posterior distribution inferred from the factual output.
The work by Bynum and Cho~\cite{bynum2024language} is fundamentally different to ours. In the context of a specific task, they utilize an SCM to describe and quantify the causal relations between variables determined by the semantics of that specific task.
In this context, it is also relevant to note that the Gumbel-Max SCM has previously been used to enable counterfactual reasoning in other domains such as Markov decision processes~\citep{tsirtsis2021counterfactual}, temporal point processes~\citep{noorbakhsh2022counterfactual}, and expert predictions~\citep{benz2022counterfactual}.

%% file: 020scm.tex
To formally express autoregressive token generation, we adopt (part of) the notation introduced by Duetting et al.~\citep{duetting2024mechanism} in a different (non-causal) context.
Let $V$ denote the vocabulary (set) of tokens available to the LLM, 
which includes an end-of-sequence token $\bot$.
Then, we denote by $V^*= V\cup V^2 \cup \dots \cup V^K$ the set of sequences of tokens up to maximum length $K$,
and by $\varnothing$ the empty token.
An LLM takes as input a prompt sequence $s_q \in V^*$ and responds with an output sequence $s \in V^*$. 
The output sequence is generated using an autoregressive process. 
At each time step $i \in [K]$, the LLM first takes as input the concatenation of the prompt sequence $s_q$ and the (partial) output sequence $s_{i-1}$ and generates a distribution over tokens $d_{i} \in \Delta(V)$. 
Then, it samples the next token $t_{i} \sim d_{i}$ from the distribution $d_{i}$ and creates the output sequence $s_{i} = s_{i-1} \circ t_{i}$, where $\circ$ denotes the concatenation of a token or sequence with another sequence. 
Further, if $t_{i} = \bot$, it terminates and returns $s = s_i$ and, otherwise, it continues to the next step $i+1$ in the generation.

\begin{figure}
    \centering
    \includegraphics[width=1\linewidth]{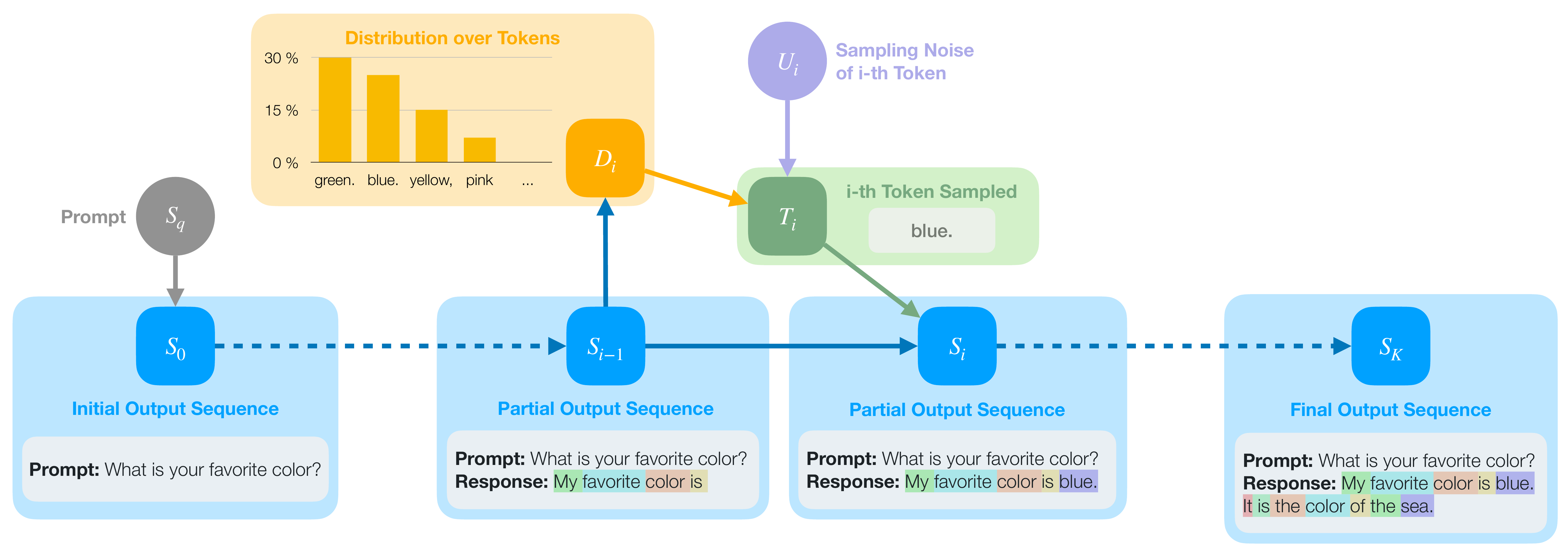}
    \caption{\textbf{Causal graph of our proposed SCM $\Mcal$ for token generation.} Boxes represent endogenous random variables and circles represent exogenous
random (noise) variables. The value of each endogenous variable
is given by a function of the values of its ancestors in the
causal graph, as defined by Eq.~\ref{eq:SCM}. The value of
each noise variable $U_i$ is sampled independently from a
given distribution $P_U$, and it determines the  stochastic state of the LLM's sampler during the generation of token $T_i$ (refer to Fig.~\ref{fig:example}).}
    \label{fig:scm}
\end{figure}

Given any prompt sequence, the above autoregressive process determines what (factual) output sequence the LLM generates as a response.
However, given a generated output sequence, the above process does not determine what counterfactual output sequence the LLM would have generated if the prompt sequence, or some of the tokens in the output sequence, had been different.
To address this limitation, we augment the autoregressive process using a structural causal model (SCM)~\citep{pearl2009causality,peters2017elements}, which we denote as $\Mcal$.
Our SCM $\Mcal$ is defined by the following assignments\footnote{We denote random variables with capital letters and realizations of random variables with lower case letters.}: 
\begin{equation}\label{eq:SCM}
\begin{split}
           S_0&=S_q,
    \quad   D_{i} 
 =\begin{cases}
        f_D(S_{i-1})
        & \text{if} \,\, \texttt{last}(S_{i-1})\neq \bot,\\
        P_\varnothing & \text{otherwise}
    \end{cases},
    \quad  T_{i} = \begin{cases}
        f_T( D_{i}, U_{i}) & \text{if} \,\, D_{i} \neq P_\varnothing,\\
        \varnothing & \text{otherwise}
    \end{cases},
        \\[2ex] 
    S_{i} &=  
    S_{i-1} \circ T_{i} 
    \quad  \text{and} \quad
    S=S_{K}, 
\end{split}
\end{equation}
where $S_q$ and $\Ub = (U_i)_{i \in \{1, \ldots, K\}}$ are independent exogenous random variables, with $S_q \sim P_{Q}$ and $U_i \sim P_{U}$, respectively,
$f_D$ and $f_T$ are given functions,
$P_\varnothing$ is the point mass distribution on $\varnothing$,
and $\texttt{last}(S_{i-1})$ denotes the last token of the sequence $S_{i-1}$.
Here, the function $f_D$ is defined by the architecture and parameters of the LLM and the choice of function $f_T$ and distribution $P_U$ determines the exact mechanism that the LLM's sampler uses to (stochastically) select the next token $T_i$. Note that,
%
there always exists a pair of $f_T$ and $P_U$ such that the distribution over tokens $D_i$ matches the distribution $P^{\Mcal}(T_i)$ entailed by $\Mcal$ (see Buesing et al.~\citep{buesing2018woulda}, Lemma 2 for a technical argument).
Moreover, note that, in the SCM $\Mcal$, the output sequence $S$ contains the prompt sequence to lighten the notation regarding interventions.
For an illustration of the SCM and its causal graph, refer to Figure~\ref{fig:scm}.

Under this augmented autoregressive process, given an output sequence $S=s$ and noise values $\Ub = \ub$, we can generate the counterfactual output sequence the LLM would have generated if the prompt sequence, or some of the tokens in the output sequence had been different, deterministically.
More formally, given an intervention $\text{do}[S_i = \tilde{s}]$, with $i \leq |s|$, the counterfactual output sequence $S = S_{K}$ can be computed recursively using the following expression:
\begin{equation}\label{eq:cf_token_selection}
    S_j = \begin{cases}
        s_j & \text{if} \,\, j<i \\
        \tilde{s} & \text{if} \,\, j=i \\
        S_{j-1} \circ f_T(f_D(S_{j-1}), u_j) & \text{if} \,\, j>i \,\, \text{and} \,\, \texttt{last}(S_{j-1})\neq \bot \\
        S_{j-1} & \text{otherwise.} 
    \end{cases}
\end{equation}

Note that the key element of this recursive expression for the counterfactual output sequence is the use of the same realized noise values $u_j$ for $j\in [K]$ that were used to generate the factual output sequence $s$.
However, without further assumptions, the counterfactual output sequence may be non-identifiable.
%
This is because there may be multiple noise distributions $P_U$ and functions $f_{T}$ under which $P^{\Mcal}(T_i) = D_i$, but  each pair produces a different counterfactual output sequence---Oberst and Sontag~\citep{oberst2019counterfactual} make a similar argument in the context of Markov decision processes.
In simpler terms, without explicitly modeling the stochastic mechanism by which the sampler selects the next token in the factual sequence, it is not possible to determine which tokens would have been selected in the counterfactual output sequence.
In the next section, we address this issue by focusing on the class of Gumbel-Max SCMs to implement an LLM's sampler.

%% file: 030implementation.tex
Under the class of Gumbel-Max SCMs, the function $f_{T}$ that implements the sampling of the next token in the SCM $\Mcal$ adopts the following functional form~\cite{oberst2019counterfactual}:
\begin{equation}\label{eq:sampling-mechanism}
    f_T( D_i, U_i) = 
        \argmax_{t \in V}\{ \log D_{i,t} + U_{i,t}\},
\end{equation}
where $U_{i,t}\sim \text{Gumbel}(0,1)$ are independently distributed Gumbel variables that determine the  stochastic state of the LLM's sampler during the generation of token $T_i$ (refer to Fig.~\ref{fig:example}).
Importantly, this class of SCMs has been shown to satisfy a desirable counterfactual stability property that can be intuitively expressed as follows. 
Assume that, at time step $i$, the augmented autoregressive process sampled token $t_i$ given $d_i=f_{D}(s_{i-1})$. 
Then, in a counterfactual scenario where $D_i=d'$, it is \emph{unlikely} that, at time step $i$, the augmented autoregressive process would have sampled a token $t'$ other than $t_i$---the factual one---unless, under the token distribution $d'$, the relative chance of generating token $t_i$ decreased compared to other tokens.
More formally, for any token distribution $d' \in \Delta(V)$ with $d' \neq d_i$ such that
\begin{equation*} 
\frac{P^{\Mcal}(T_i = t_i \given D_i = d')}{P^{\Mcal}(T_i = t_i \given D_i = d_i)}\geq
\frac{P^{\Mcal}(T_i = t' \given D_i = d')}{P^{\Mcal}(T_i = t' \given D_i = d_i)},
\end{equation*}
it holds that, in the counterfactual scenario where $D_i=d'$, the counterfactual token $T_i \neq t'$.
%

To understand the intuition behind counterfactual stability, consider the following example.
Assume that the vocabulary $V$ contains 2 tokens ``A'' and ``B'', the (factual) distribution $d_i$ assigns values $0.6$ and $0.4$, respectively, and the Gumbel-Max SCM samples token ``A''. Consider an intervention that changes $d_i$ to a distribution $d'$ that assigns values $0.7$ and $0.3$ to ``A'' and ``B'', respectively. 
Then, the property of counterfactual stability ensures that, during the counterfactual generation, token ``A'' is also sampled because, in comparison to the factual generation, the relative odds are higher, \ie, $0.7$ to $0.3$ vs. $0.6$ to $0.4$.
In a way, counterfactual stability ``prioritizes'' the token sampled in the factual generation, 
maintaining consistency between the factual and counterfactual text.
For a further discussion of the stability property and alternatives to the Gumbel-Max SCM, refer to Section~\ref{sec:discussion}.

\SetKwComment{Comment}{/* }{ */}

\begin{algorithm}[t!]
  \textbf{Input}: Random number generator states $\rb$, factual output sequence $s$, intervention $(i, \tilde{s})$. \\
  \textbf{Output}: Counterfactual output sequence $s'$. \\
  \For{$j=1,\ldots,K$}{
	  \If{$j < i$} {
            $s'_{j} = s_j$
        }
        \ElseIf{$j = i$} {
            $s'_{j} = \tilde{s}$
        }
        \ElseIf{$j > i \wedge \texttt{last}(s'_{j-1}) \neq \bot$} {
            $u_j = \text{GenGumbel}(r_j)$ \\
            $d'_{j,t} = f_{D}(s'_{j-1})$ \\
            $t_j = \argmax_{t \in V} \{ \log d'_{j,t} + u_{j,t} \}$ \\
            $s'_{j} = s'_{j-1} \circ t_j$
        }
        \Else {
            $s'_j = s'_{j-1}$
        }
  }
  \textbf{Return} $s'_K$  
  \caption{It returns a counterfactual sequence of tokens using a Gumbel-Max SCM} \label{alg:counterfactual-generation}
  \end{algorithm}
  
In practice, in addition to solving the non-identifiability issues discussed previously, the use of Gumbel-Max SCMs allows for an efficient procedure to sample a sequence of counterfactual tokens with minimal additional memory requirements compared to vanilla token generation.
We summarize the procedure in Algorithm~\ref{alg:counterfactual-generation}.
Recall that, to generate the counterfactual output sequence, one needs to use the same values $u_j$ for the noise variables that were used during the factual generation and then perform an autoregressive computation based on Equation~\ref{eq:cf_token_selection}.
Instead of storing the values $u_j$  for all time steps $j\in [K]$, whose dimensionality matches the size of the vocabulary $V$, Algorithm~\ref{alg:counterfactual-generation} employs a simple idea: it stores the state of the random number generator $r_j$ used at each time step $j\in[K]$ of the factual generation. Then, during the counterfactual generation, it regenerates the values $u_j=\text{GenGumbel}(r_j)$  on the fly.\footnote{Storing the realized values of the Gumbel variables requires storing $\mathcal{O}(KV)$ float values since $u_j \in \mathbb{R}^V$. On the other hand, the states of random number generators $r_j$ take values in $\mathbb{N}^d$, where, for instance, $d = 16$ in \texttt{pytorch}~\citep{paszke2019pytorch}. Thus, our approach requires $\mathcal{O}(K)$ additional integer memory compared to vanilla token generation.}
%

\xhdr{Remarks on implementation aspects of LLMs} In practice, to avoid sampling tokens with very low probability, LLMs may not directly sample from the distribution over tokens $d_i$ at each time step $i$. 
Instead, a common practice is to sample from a distribution $\hat{d}_{i} \in \Delta(V_i)$, where $\hat{d}_{i,t} \propto d_{i,t}$ if $t \in V_i$ and $\hat{d}_{i,t} = 0$ otherwise, where $V_i$ is either the set of most likely tokens of size $k$ under $d_i$---known as ``top-$k$'' sampling---or the set of most likely tokens whose cumulative probability exceeds a given value $p$ under $d_i$---known as ``top-$p$'' or ``nucleus'' sampling~\citep{holtzman2019curious}. 
We can readily implement top-$k$ sampling and top-$p$ sampling in the SCM $\Mcal$ by restricting the $\argmax$ in Equation~\ref{eq:sampling-mechanism} to the respective set $V_i$. However, in general, the resulting model is not guaranteed to satisfy counterfactual stability.

In all state-of-the-art LLMs, to ensure that the distribution $d_i$ over tokens at each time step $i$ is a valid probability distribution, the final layer in their neural network is a softmax layer. A crucial feature of this layer is the \emph{temperature} parameter, $\tau$, which controls the level of uncertainty in $d_i$. Intuitively, higher values of $\tau$ result in a more uniform distribution, while as $\tau$ approaches zero, the distribution concentrates increasingly on the most probable next token. In the next section, we perform a series of experiments in which we analyze the performance of counterfactual token generation, examining the effects of varying temperature values, as well as the application of top-$k$ and top-$p$ sampling.

%% file: 040experiments.tex
In this section, we experiment with an implementation of our model on \texttt{Llama 3 8B-Instruct}~\citep{dubey2024llama} and \texttt{Ministral-8B-Instruct}~\citep{jiang2023mistral}, two popular open-weights large language models.
We start by qualitatively analyzing an example of counterfactual story generation. Next, we quantitatively analyze the similarity between factual and counterfactual text.
We conclude with an application of counterfactual token generation in detecting model biases towards demographic groups.\footnote{All experiments ran on an 
cluster of machines, each equipped with 24 Intel(R) Xeon(R) 3GHz CPU cores, 1024GBs of memory and 2 NVIDIA A100 80GB GPUs.}

\subsection{How would the story have unfolded for ``Captain Maeve''?\label{sec:stories}}

As discussed in Section~\ref{sec:counterfactual}, by using the Gumbel-Max SCM, our approach to counterfactual token generation is guaranteed to satisfy the property of counterfactual stability---counterfactual token generation ``prioritizes'' selecting the same tokens $T_i$ that were selected during the factual generation.
As a consequence, we expect the counterfactual text generated using counterfactual token generation to be similar to the factual text. Here, we investigate this qualitatively through an anecdotal example of story generation.

\begin{figure}[t]
    \centering
    \subcaptionbox{Factual story\label{fig:stories-f}}{
    \includegraphics[width=0.9\linewidth]{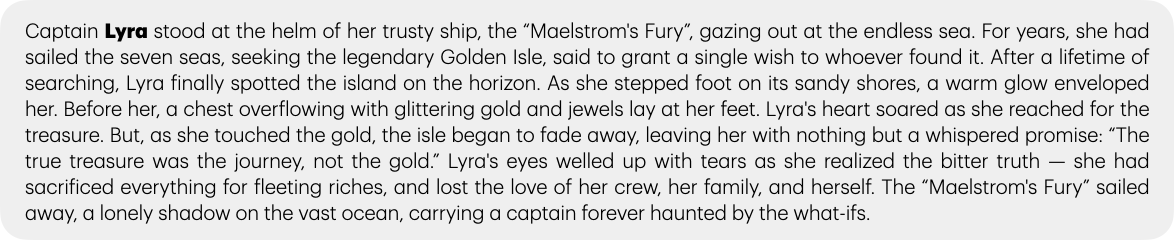}}
    \\
    \subcaptionbox{Story generated with interventional token generation, after modifying the protagonist's name \label{fig:stories-int}}{
    \includegraphics[width=0.9\linewidth]{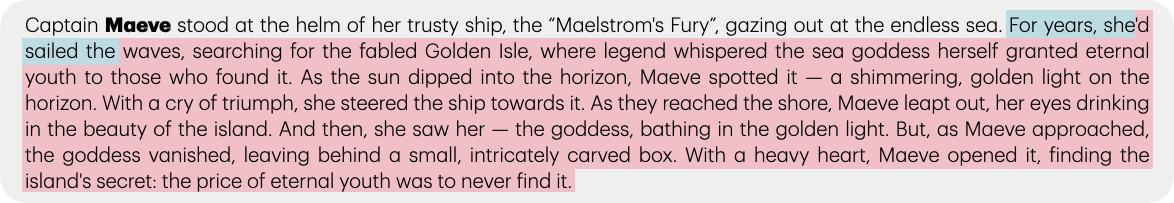}}
    \\
  \subcaptionbox{Story generated with counterfactual token generation, after modifying the protagonist's name\label{fig:stories-cf}}{
    \includegraphics[width=0.9\linewidth]{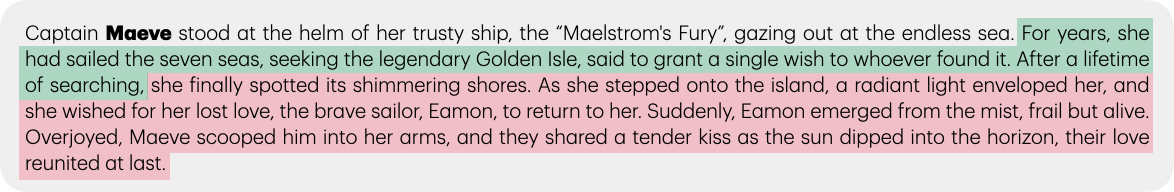}}
    \caption{\textbf{Examples of factual, interventional and counterfactual stories. }Panel (a) shows a factual story, as given by \texttt{Llama 3 8B-Instruct}. Panels (b) and (c) show variants of the story resulting from~in\-ter\-ven\-tional and counterfactual token generation, respectively. In panels (b), (c), we give as input to the LLM the original prompt along with the first sentence of the factual output (non-highlighted text), modified by replacing ``Lyra'' with ``Maeve''. Blue (green)-highlighted text indicates the tokens of the output that are identical in the factual story and its interventional (counterfactual) counterpart.
    Red-highlighted text indicates the differences. In both panels, the temperature parameter is set to $\tau=0.9$.}
    \label{fig:stories}
\end{figure}

We use the implementation of our model on \texttt{Llama 3 8B-Instruct} with the system prompt \textit{``Be creative and keep your response as short as possible.''} and a query prompt \textit{``Tell me a fantasy story about a captain. The story should have either a happy or a sad ending.''} Figure~\ref{fig:stories-f} shows the (factual) generated story about Captain Lyra, her ship the Maelstrom's Fury, and her quest to find a treasure on the Golden Isle. Then, we use the original prompt along with part of the factual output (\ie, the first sentence of the story) as input to the model, modifying the protagonist's name from ``Lyra'' to ``Maeve'', and we regenerate the rest of the output using two approaches:
\begin{enumerate}
    \setlength\itemsep{0em}
    \setlength\parskip{0em}
    \item \textbf{Interventional token generation (see Fig~\ref{fig:example-int}):} it regenerates the second part of the output using vanilla autoregressive token generation, \ie, it samples new noise values $u_j$ for the second part of the output using different states $r_j$ for the random number generator from those used in the factual generation.
    \item \textbf{Counterfactual token generation (see Fig~\ref{fig:example-cf}):} it regenerates the second part of the output using Algorithm~\ref{alg:counterfactual-generation}, \ie, it reuses the same states $r_j$ for the random number generator and, hence, the same noise values $u_j$ as the ones used in the factual generation.
\end{enumerate}

Figures~\ref{fig:stories-int},~\ref{fig:stories-cf} present two alternative versions of the factual story generated using the methods mentioned above. These stories reveal several interesting insights. The story generated with interventional token generation starts diverging from the factual story after only a few tokens, as the method lacks memory of the noise values $u_j$ that resulted in the original output. In contrast, the initial part of the counterfactual output remains identical to the factual output, as expected, due to the counterfactual stability property of the Gumbel-Max SCM and the minor nature of changing the protagonist's name.
Although one may expect this to apply for the rest of the counterfactual output, thinking that the protagonist's name would be irrelevant to the narrative of this particular story, this is not the case. 
Perhaps surprisingly, the use of ``Maeve'' instead of ``Lyra'' results in a partially different output, 
illustrating that the LLM's probability distributions over next tokens are sensitive even to minor changes. In Appendix~\ref{app:stories}, we also observe differences between the factual and counterfactual outputs resulting from other seemingly irrelevant interventions, such as changing the name of the ship, removing the adjective ``trusty'' or replacing the word ``sea'' with ``blue''.

\begin{figure}[t]
    \centering
    \subcaptionbox*{}{
    \centering
    \includegraphics[width=0.8\textwidth]{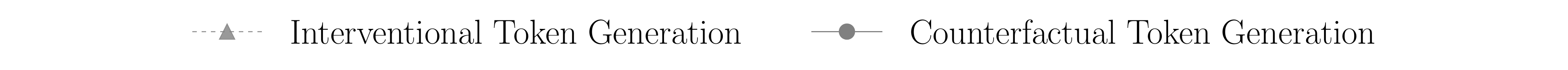}
    \vspace{-6mm}
    }
    \subcaptionbox{Gumbel-Max SCM~\label{fig:temp-vs-ed}}{
        \centering
        \includegraphics[width=0.3\textwidth]{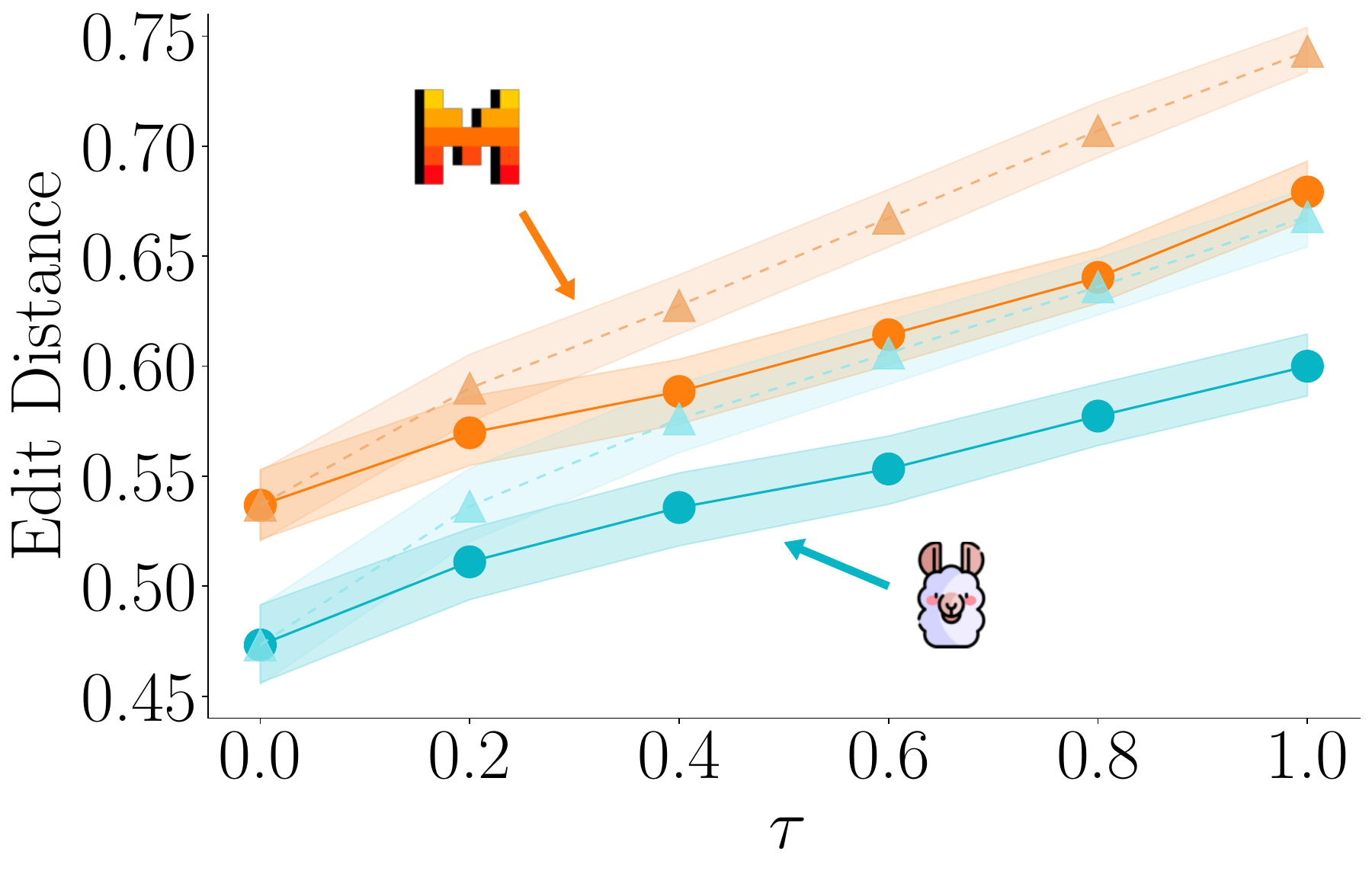}
    }
    \subcaptionbox{Top-$p$ Gumbel-Max SCM~\label{fig:p-vs-ed}}{
        \centering
        \includegraphics[width=0.3\textwidth]{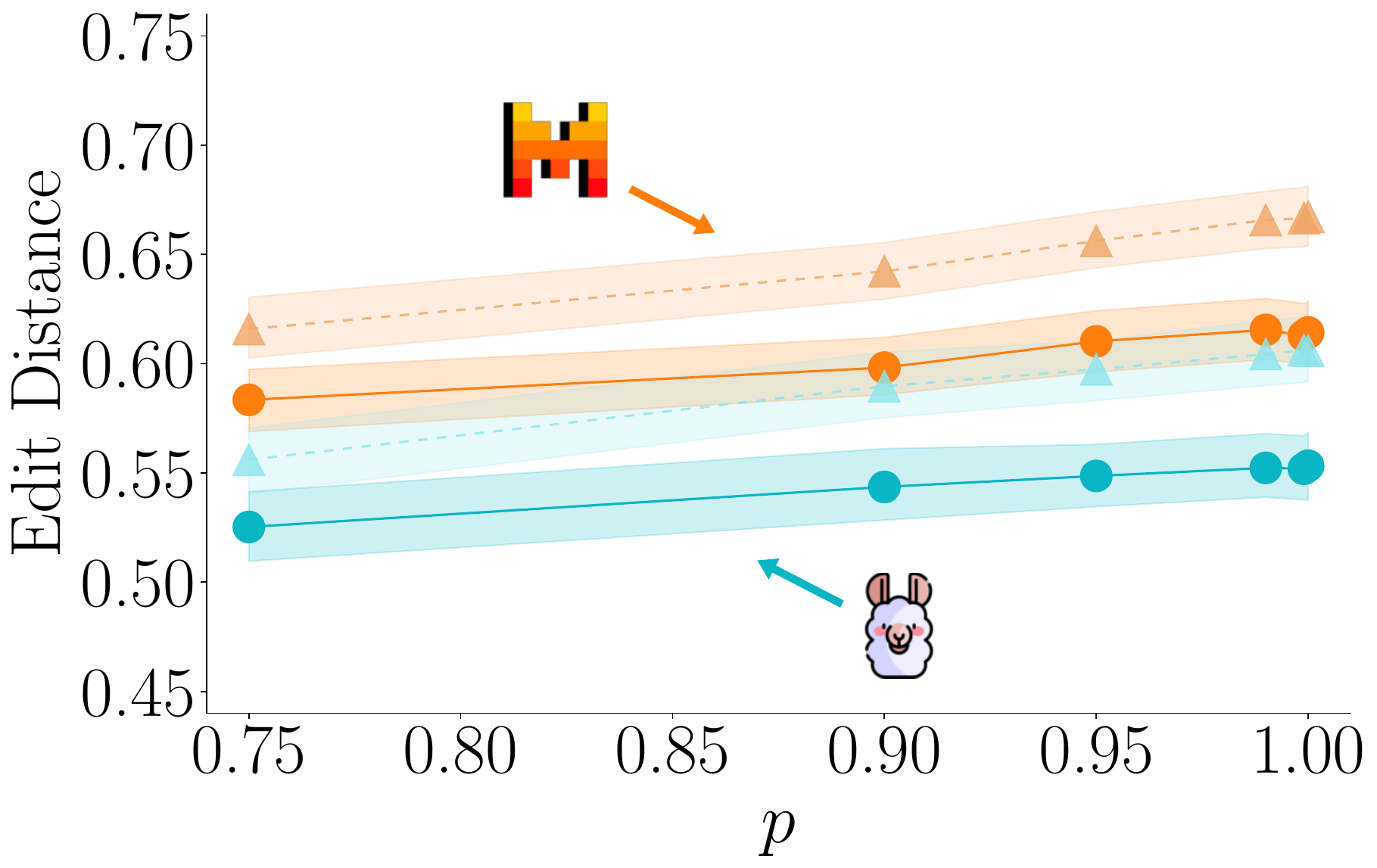}
    }
    \subcaptionbox{Top-$k$ Gumbel-Max SCM~\label{fig:k-vs-ed}}{
        \centering
        \includegraphics[width=0.3\textwidth]{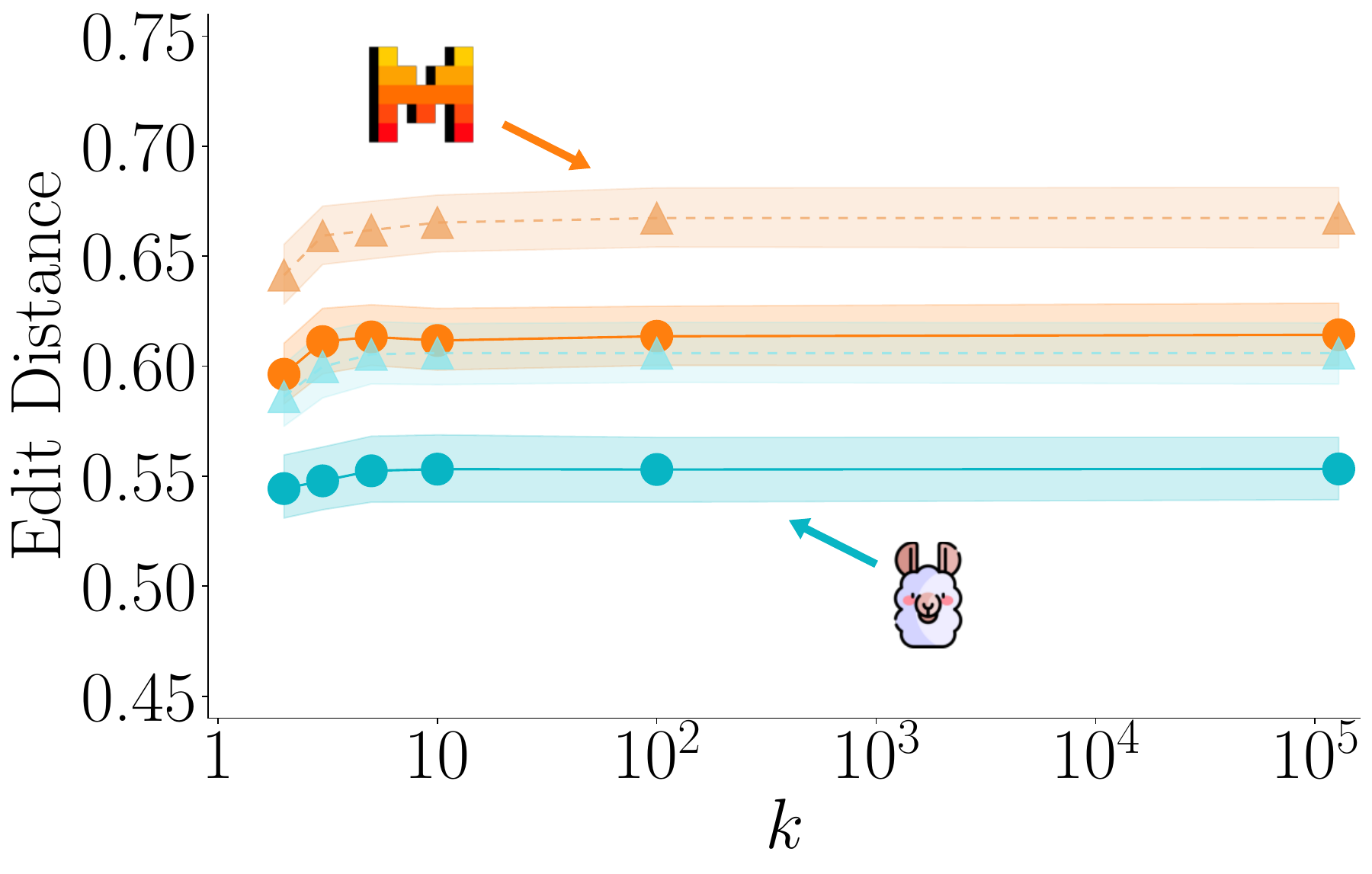}
    }
    \caption{\textbf{Comparison between interventional and counterfactual token generation.} The panels show the edit distance between the factual token sequence and the sequence generated by interventional and counterfactual token generation using (a) the Gumbel-Max SCM defined in Equation~\ref{eq:sampling-mechanism}, (b) the top-$p$ Gumbel-Max SCM, and (c) the top-$k$ Gumbel-Max SCM discussed at the end of Section~\ref{sec:counterfactual}, against various values of the temperature parameter $\tau$, $p$ and $k$, respectively.
    In panels (b, c) the temperature parameter is set to $\tau = 0.6$.
    In all three panels, the edit distance is averaged over $4{,}000$ output sequences, resulting from two independent interventions per factual sequence, and shaded areas represent $95\%$ confidence intervals. 
    The icons \includegraphics[height=0.99\inlineheight]{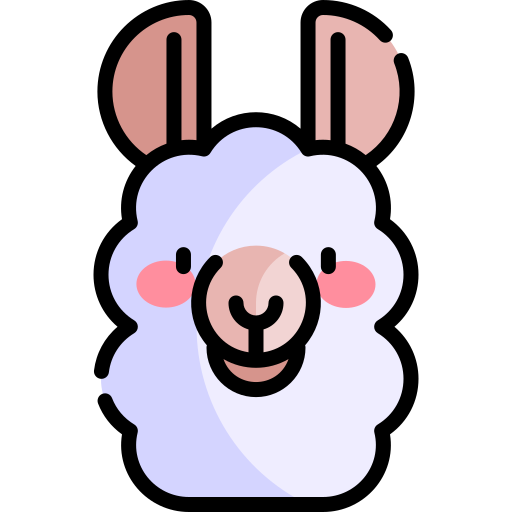} and \includegraphics[height=0.9\inlineheight]{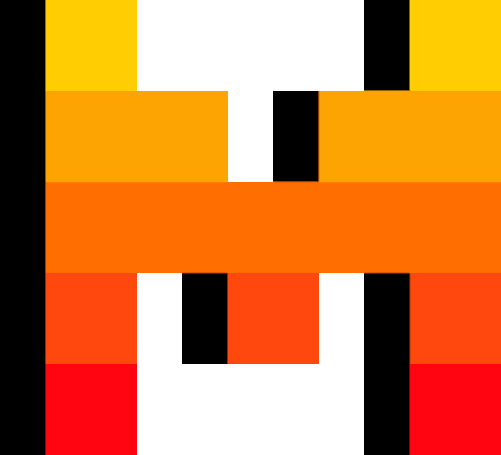} indicate results for \texttt{Llama 3 8B-Instruct} and \texttt{Ministral-8B-Instruct} respectively.
    }
    \label{fig:edit-distance-vs-params}
\end{figure}

\subsection{How similar is counterfactually generated text to the factual one?}\label{sec:similarity}

In the previous section, we demonstrated through an example that counterfactual token generation results in text that is (partially) similar to the factual text, as expected due to the property of counterfactual stability. Here, we empirically verify this expectation using a quantitative analysis and explore how it is affected by the model parameters.

\xhdr{Experimental setup} We first use the implementation of our model on \texttt{Llama 3 8B-Instruct} and \texttt{Ministral-8B-Instruct} to generate (factual) outputs to $2{,}000$ question prompts sourced from the LMSYS Chat 1M dataset~\cite{zheng2024lmsyschat}. As a system prompt we use \textit{``Keep your replies short and to the point.''}.
Further, for each factual output, we perform two interventions where we replace a randomly selected token $t_i$ with a token $t' \neq t_i$.\footnote{To select $t'$, we set the probability of $t_i$ in $d_i$ to $0$, re-scale the values of $d_i$  and use top-$p$ sampling with $p=0.9$.} 
One of the two interventions restricts the choice of $t_i$ to the first half of the output sequence and the other restricts it to the second half. 
Then, for each intervened factual output, we feed the concatenation of the question prompt and the first part of the intervened factual output up to and including token $t'$ as input to our model. We regenerate the second part of the output after token $t'$ using (i) interventional token generation and (ii) counterfactual token generation, as described in Section~\ref{sec:stories}.
Finally, we measure the lexicographic similarity between the regenerated second part of the output and its factual counterpart using their (normalized) Levenshtein edit distance~\cite{levenshtein1966binary}.
In our experiments, we implement our model using the Gumbel-Max SCM defined in Equation~\ref{eq:sampling-mechanism} as well as the top-$p$ Gumbel-Max SCM and top-$k$ Gumbel-Max SCM discussed at the end of Section~\ref{sec:counterfactual}.

\xhdr{Results} Figure~\ref{fig:edit-distance-vs-params} summarizes the results, which show that, across different SCMs and LLMs, the output sequences generated using counterfactual token generation are more similar to the factual sequences (\ie, the edit distance is lower) than the output sequences generated using interventional token generation.
This suggests that, even though the top-$p$ and top-$k$ Gumbel-Max SCMs are not guaranteed to satisfy counterfactual stability, in practice, counterfactual token generation under both SCMs does ``prioritize'' selecting the same tokens $T_i$ that were selected during the factual generation.
Although this pattern is consistent across the two LLMs, we observe that output sequences generated by both interventional and counterfactual token generation are more similar to their respective factual sequences for \texttt{Llama 3 8B-Instruct} than \texttt{Ministral-8B-Instruct}.

\begin{figure}[t]
    \captionsetup[subfigure]{justification=centering}
    \centering
    \subcaptionbox{Change in income upon intervention on sex (direct effect)~\label{fig:sex-income-direct}}{
        \includegraphics[width=0.285\textwidth]{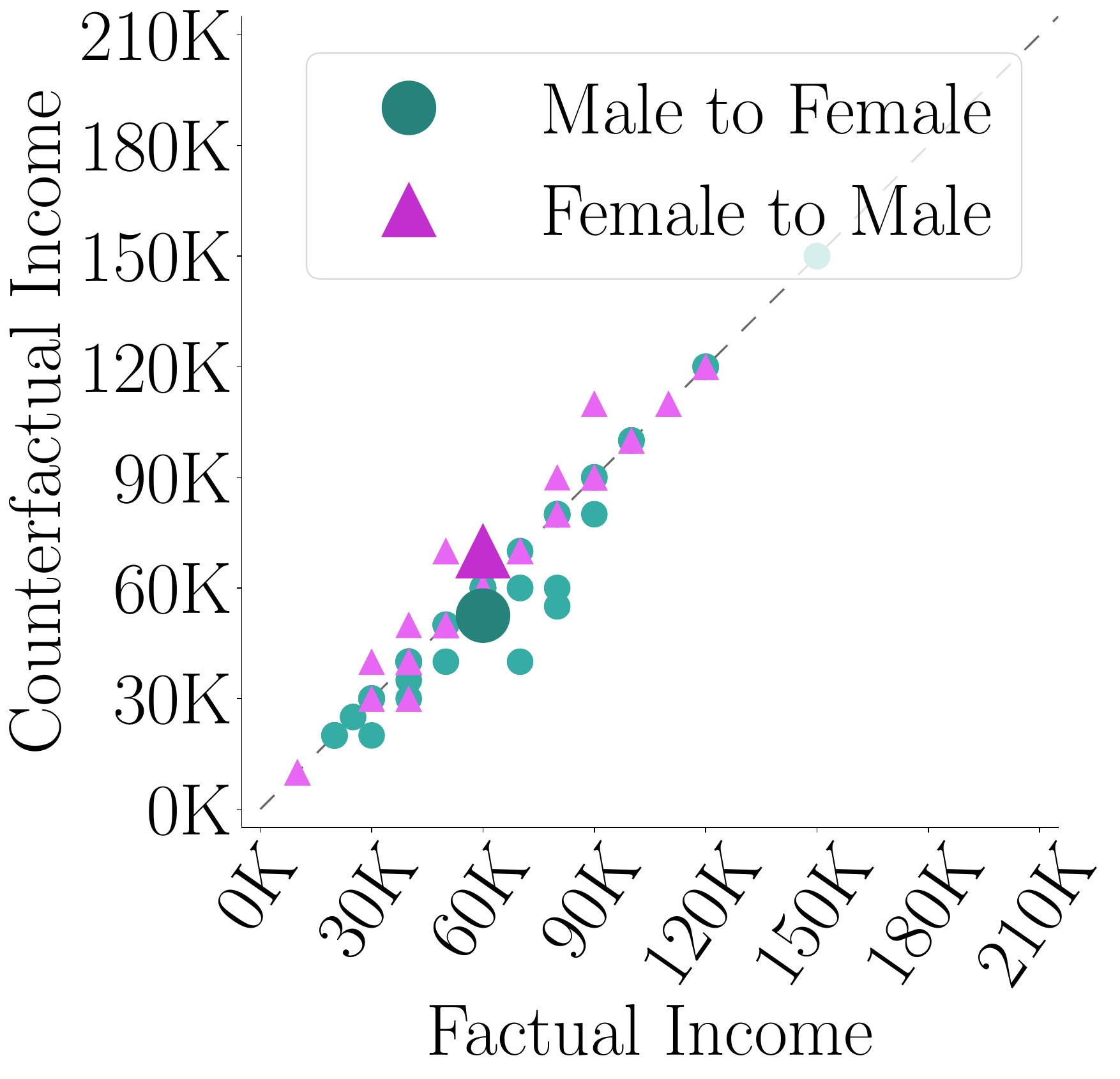}
    }
    \subcaptionbox{Change in income upon intervention on sex (total effect)~\label{fig:sex-income-total}}{
        \includegraphics[width=0.285\textwidth]{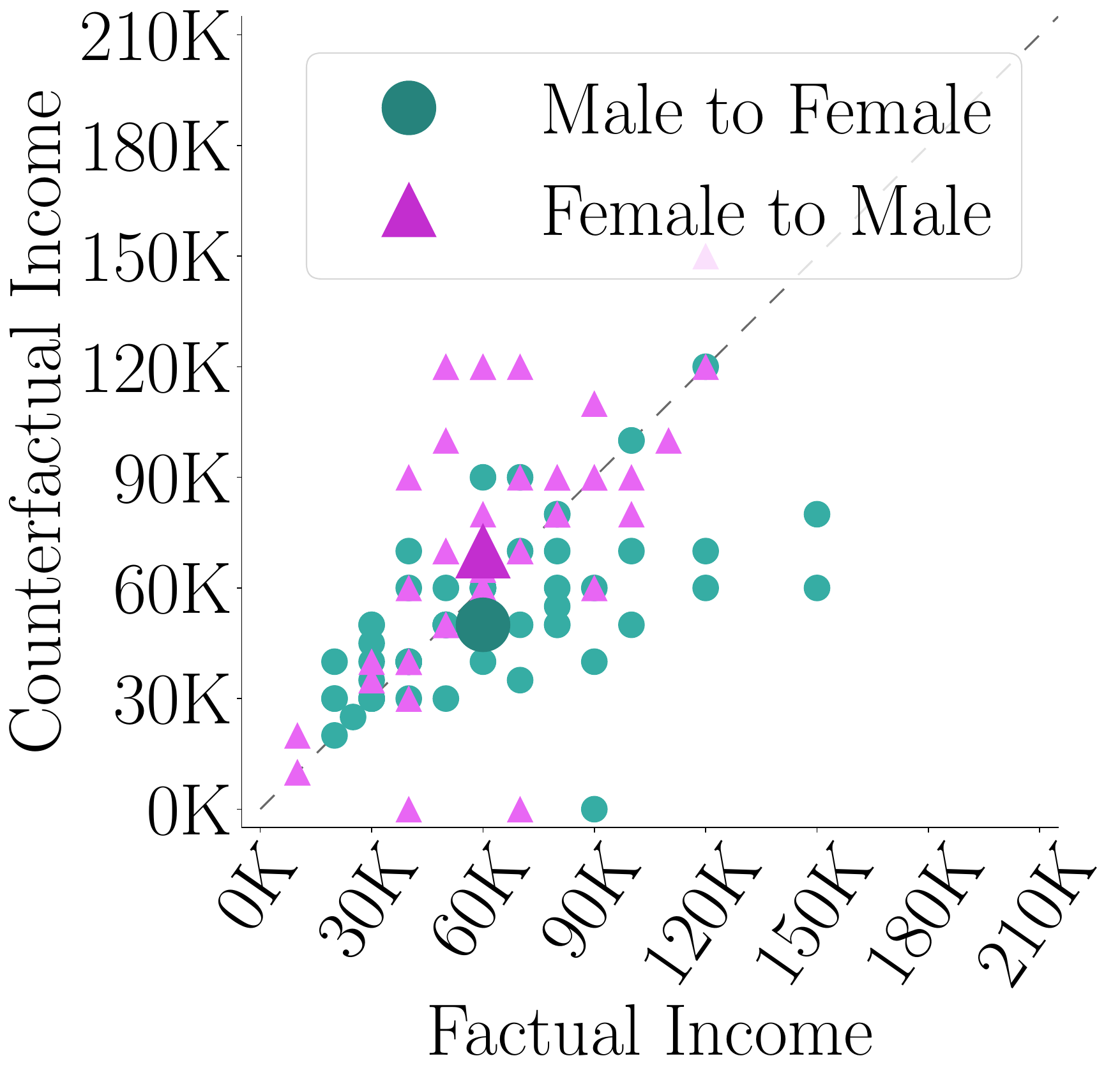}
    }
    \subcaptionbox{Distribution of factual\\and counterfactual income~\label{fig:sex-income-hist-fem}}{
        \includegraphics[width=0.335\textwidth]{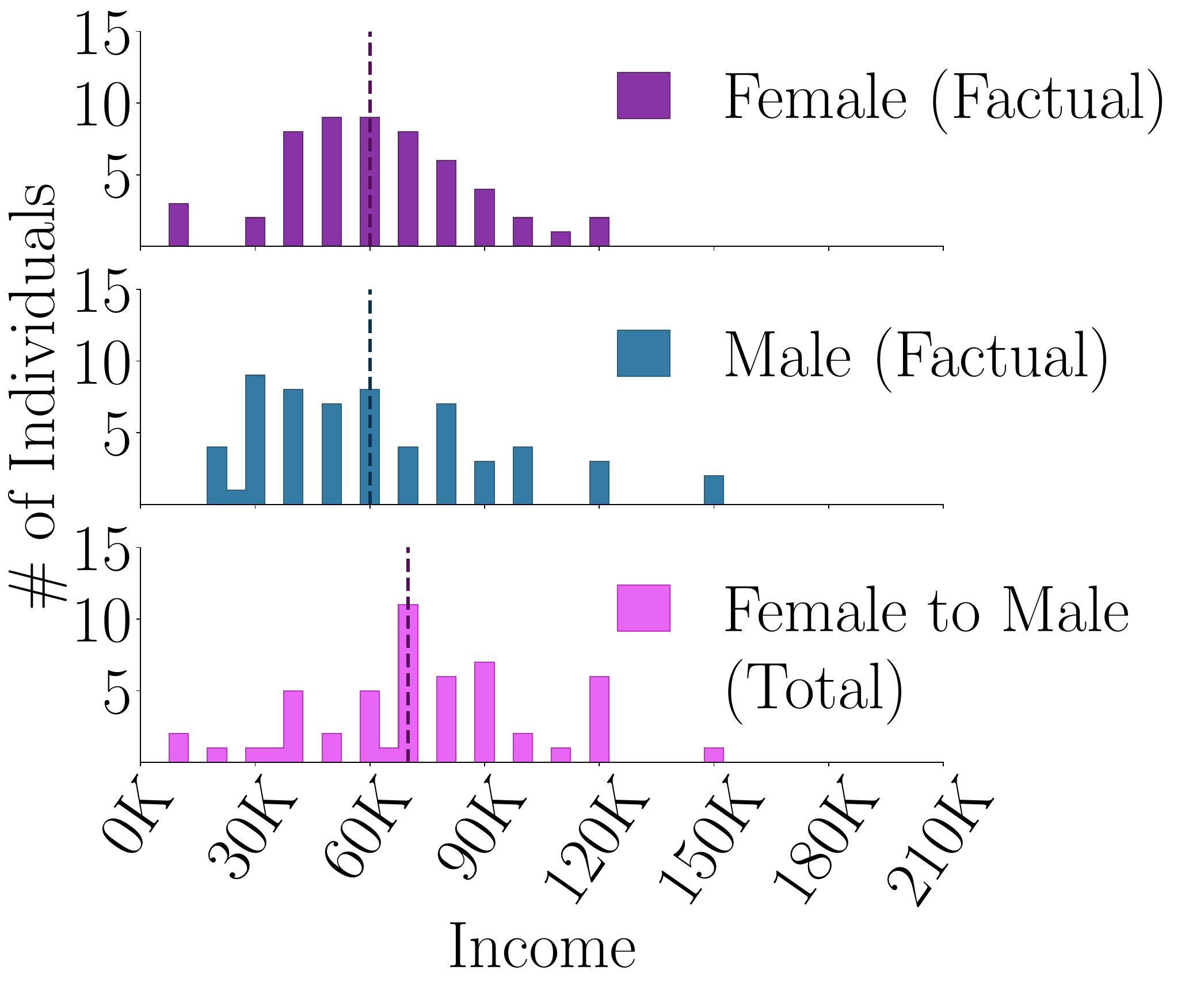}
    }
    
    \caption{
    \textbf{Comparison between factual and counterfactual income.}
    Panel (a) shows the change in income of male (female) individuals had they been female (male), while keeping fixed the rest of their attributes preceding income in the output sequence.
    Panel (b) shows the change in income of male (female) individuals had they been female (male), while keeping fixed the attributes preceding sex but allowing the attributes between sex and income to change in the output sequence.
    %
    Panel (c) shows the factual distributions of income of female and male individuals and the counterfactual distribution of income of female individuals under the same intervention as in panel (b).
    Enlarged points in panels (a, b) and dashed lines in panel (c) correspond to the median income.
    In all panels, we use \texttt{Llama 3 8B-Instruct} and set the temperature parameter to $\tau=0.8$.
    }
    \label{fig:bias-income}
\end{figure}

\subsection{Does counterfactual token generation reveal model biases?}\label{sec:bias}
Common approaches to addressing questions of bias and fairness rely on making counterfactual comparisons based on sensitive attributes~\citep{kusner2017counterfactual}. For example, would a person'{}s income have been the same if their race or sex were different? In this section, we focus on a census data generation task, and demonstrate the use of counterfactual token generation to investigate potential biases of an LLM towards demographic groups.

\xhdr{Experimental setup}
We first use the implementation of our model on \texttt{Llama 3 8B-Instruct} and \texttt{Ministral-8B-Instruct} to generate (factual) census data. For each model, we use the same input prompt three times with different seeds (see Appendix~\ref{app:bias-setup} for details), requesting $50$ individuals each time. The attributes of generated individuals include their age, sex, citizenship, race, ethnicity, marital status, number of children, occupation, income, and education, in this given order, and we discard individuals for whom the generated income is exactly zero. The factual data generated by the models result in a corpus of $114$ and $158$ fictional individuals for \texttt{Llama 3 8B-Instruct} and \texttt{Ministral-8B-Instruct}, respectively. 

For each fictional person, we consider all possible interventions on each of the sensitive attributes of sex and race.
For each intervention on sex and race, we concatenate the input prompt with the initial part of the output that includes the fictional person'{}s description up to and including the intervened sensitive attribute.
This concatenated input is then used by our model to regenerate the latter part of the output, following the intervention, using counterfactual token generation (\ie, Algorithm~\ref{alg:counterfactual-generation}).
Then, we compare the factual and counterfactual values of income, education and occupation to measure the \emph{total effect} of sex and race on those attributes.
In addition, for each intervention on sex, we also measure the \emph{direct effect} of sex on income. To this end, we concatenate the input prompt with the initial part of the output that includes the intervened value of sex and the factual values of all other attributes preceding income. 
This concatenated input is then used by our model to regenerate the latter part of the output starting from the income attribute.\footnote{For further details regarding the difference between total and direct effects, refer to Pearl~\citep{pearl2009causality}, Chapter 4.}

\begin{figure}
    \captionsetup[subfigure]{justification=centering}
    \centering
    \subcaptionbox{Change in education level\\upon intervention on race~\label{fig:race-education}}{
        \includegraphics[width=0.285\textwidth]{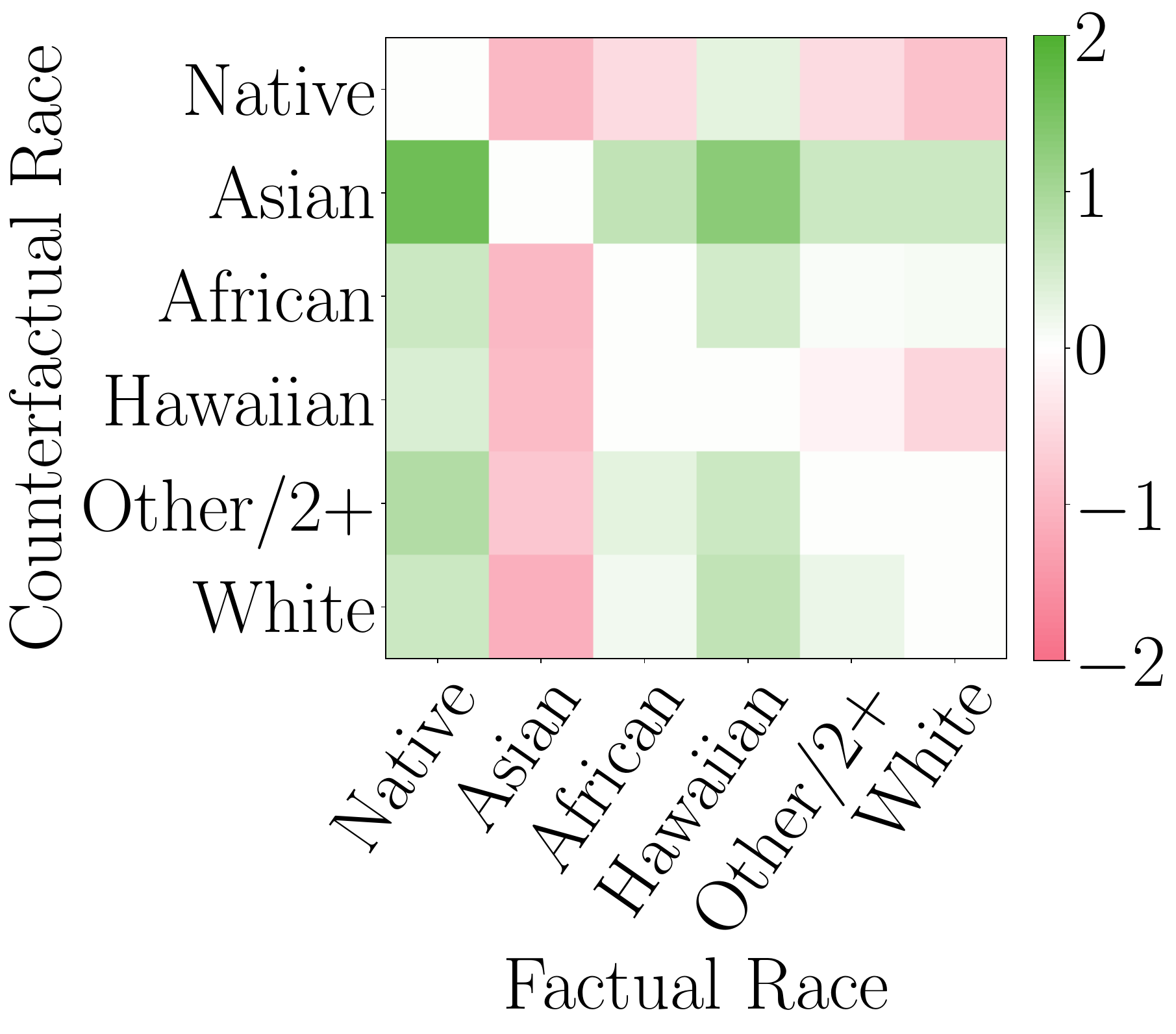}
    }
    \subcaptionbox{Change in occupation\\upon intervention on race~\label{fig:race_occupation}}{
        \includegraphics[width=0.37\textwidth]{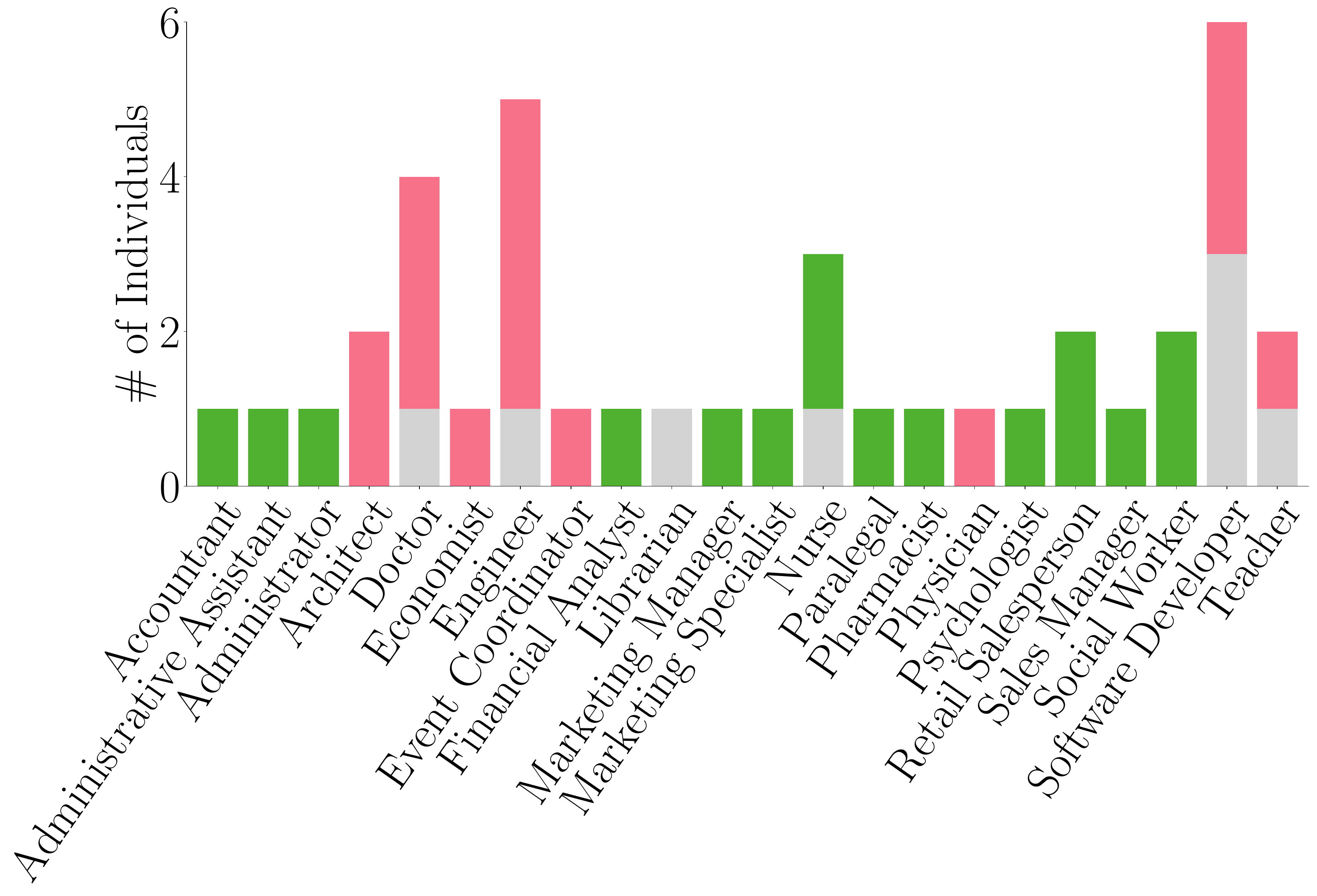}
    }
    
    \caption{
    \textbf{Comparison between factual and counterfactual education and occupation.}
    Panel (a) shows the average difference in the education level of individuals of each race had their race been different.
    Here, positive values indicate an improvement in education and negative values indicate a decline.
    Panel (b) shows the distribution shift of occupations among Asian American individuals had they been Black or African American. 
    Green (red) sections indicate the counterfactual increase (decrease) in the number of individuals that practice each occupation.
    In both panels, we use \texttt{Ministral-8B-Instruct} and set the temperature parameter to $\tau=0.8$.}
    \label{fig:bias-education-occupation}
\end{figure}
    
\xhdr{Results}
Figure~\ref{fig:bias-income} summarizes the results with respect to the effect of sex on income using \texttt{Llama 3 8B-Instruct}. 
%
For both the direct and the total effect of sex on income, Figures~\ref{fig:sex-income-direct} and ~\ref{fig:sex-income-total} show that the generated income for most male (female) individuals would have remained unchanged or decreased (increased) had they been female (male), however, the total effect exhibits larger variance.
%
%
This suggests that, in the LLM's world model, both the direct and total effects of sex on income are present,
but at a moderate level.
%
%
%
Interestingly, the (potentially biased) effect of sex on income cannot be identified solely from the factual distributions of female and male income, which present exactly the same median, as shown in Figure~\ref{fig:sex-income-hist-fem}.
%
For additional results using \texttt{Llama 3 8B-Instruct} and \texttt{Ministral-8B-Instruct}, refer to Appendix~\ref{app:bias-plots}.

%
%

%
%
Figure~\ref{fig:bias-education-occupation} summarizes the results with respect to the effect of race on education and occupation using \texttt{Ministral-8B-Instruct}.
Figure~\ref{fig:race-education} shows that, for individuals of all (generated) races, there exists at least one other race that, had they belonged to it, they would have experienced a significant increase or decrease in their education level (refer to Appendix~\ref{app:bias-setup} for the assignment of each education level to a numerical value and for the factual distribution of individuals across races).
Specifically, we observe a consistent pattern in which the LLM would have increased the level of education for (i) American Indian or Alaska Native (Native) and (ii) Native Hawaiian or Other Pacific Islander (Hawaiian) individuals had they belonged to any other race. In contrast, it would have decreased the level of education for Asian American (Asian) individuals.\footnote{The terms for races in parentheses and in Figure~\ref{fig:race-education} are shortened descriptions, which we use for brevity. Refer to Appendix~\ref{app:bias-setup} for a list of all official descriptions.}
%
Figure~\ref{fig:race_occupation} shows that, for Asian American individuals, their occupation would have shifted from STEM to humanities-related occupations had they been Black or African American. In Appendix~\ref{app:bias-plots}, we present qualitatively similar results using \texttt{Llama 3 8B-Instruct}.

%% file: 050discussion.tex
In this section, we discuss several assumptions and limitations of our work, pointing out avenues for future research. 

\xhdr{Methodology}
Our causal model of the autoregressive process underpinning large language models operationalizes the counterfactual stability property using the Gumbel-Max SCM. 
It would be interesting to understand the sensitivity of counterfactual token generation to this specific 
choice of SCM and implement counterfactual token generation using alternative SCMs obeying counterfactual stability~\cite{lorberbom2021learning, haugh2023counterfactual}.
In this context, it is also important to acknowledge that, while counterfactual stability is somewhat an appealing property, Haugh and Singal~\cite{haugh2023counterfactual} have recently argued that its appropriateness depends on the application and should be justified by domain specific knowledge. 
Moreover, they have shown that there are cases where counterfactual stability may permit counterfactuals that it was designed to exclude.
Motivated by this observation, in Appendix~\ref{app:sampler}, we include additional experiments on counterfactual token generation using a classical sampler for categorical distributions, which can be viewed as an SCM that does not satisfy counterfactual stability.

Further, there are reasons to believe that counterfactual statements generated using our model may be inconsistent with the underlying causal model of the world shared by humans~\citep{li2021implicit, li2022emergent, patel2022mapping,vafa2024evaluating} and this may render them unreliable for certain uses cases.
An interesting future direction would be to explore the use of our methodology in conjunction with human feedback to train (or fine-tune) LLMs that better understand causal relationships.

\xhdr{Applications}
%
Counterfactual token generation may be useful for understanding the inner workings of an LLM.
For example, it may be used as a tool for 
evaluating causal dependencies among physical or social attributes within the world model learned by an LLM, as in our bias experiments in Section~\ref{sec:bias}, or quantifying the ``importance'' of different parts of a model's output in reaching a final conclusion, similarly as in feature attribution~\citep{ribeiro2016should,adebayo2018sanity}.
Further, counterfactual token generation may also be useful for creating novel interfaces for LLM-human collaborations.
For example, it could be used in scenarios in which a user is satisfied with a portion of an LLM's output but wishes to modify a few words while maintaining similarity between the subsequent text and the original output. 
Additionally, counterfactual token generation may be proven useful in (applications in) linguistics and cognitive science.



\xhdr{Evaluation}
We have implemented our model on two LLMs, namely \texttt{Llama 3 8B-Instruct} and \texttt{Mi\-ni\-stral-8B-Instruct}. It would be useful to implement our model on other LLMs and use counterfactual token generation to reveal similarities and differences between the underlying models of the world constructed by different LLMs. In this context, it would be insightful to see whether the sensitivity of an LLM's counterfactual output changes as its number of parameters increases.

%% file: 060conclusions.tex
In this work, we have proposed a causal model of token generation. Using the Gumbel-Max SCM, we have introduced a methodology that enhances state-of-the-art LLMs with the ability to perform counterfactual token generation, allowing them to reason about past alternatives to their own outputs. We have experimentally analyzed the similarity between an LLM's original output and the one generated by counterfactual token generation, and we have demonstrated the use of our methodology in bias detection.

%% file: 070appendix.tex
\section{Additional counterfactual stories}\label{app:stories}


\begin{figure}[h!]
\centering
    \subcaptionbox{Factual story}{
    \includegraphics[width=0.9\linewidth]{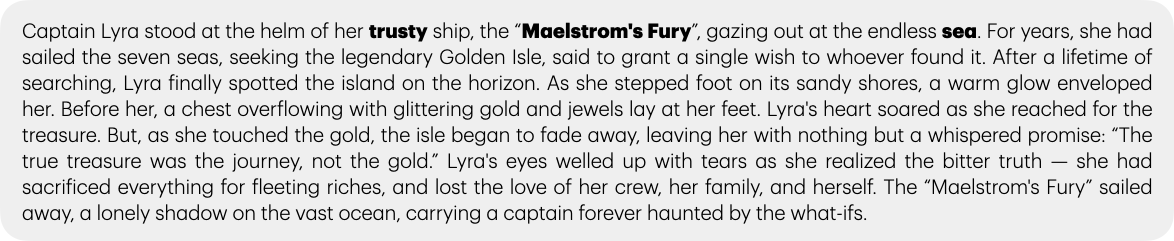}}
    \\
  \subcaptionbox{Story generated with counterfactual token generation, after modifying the ship's name}{
    \includegraphics[width=0.9\linewidth]{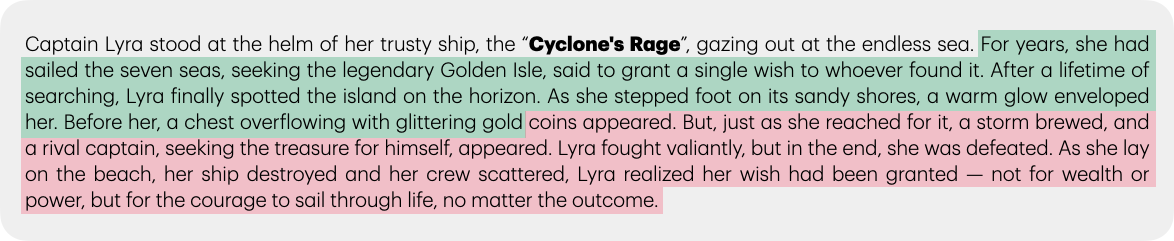}}
    \\
  \subcaptionbox{Story generated with counterfactual token generation, after changing ``sea'' to ``blue''}{
    \includegraphics[width=0.9\linewidth]{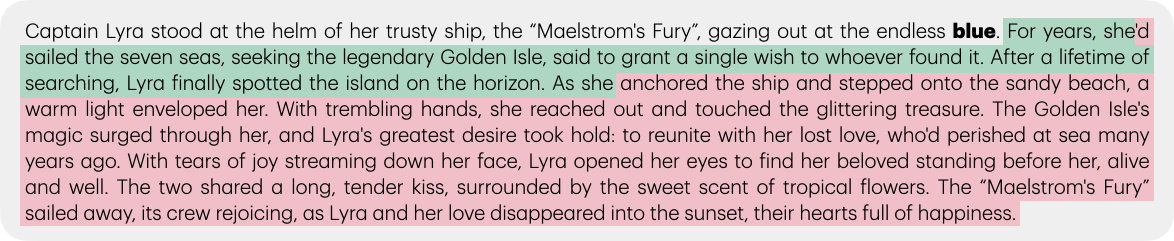}}
    \\
  \subcaptionbox{Story generated with counterfactual token generation, after deleting the word ``trusty''}{
    \includegraphics[width=0.9\linewidth]{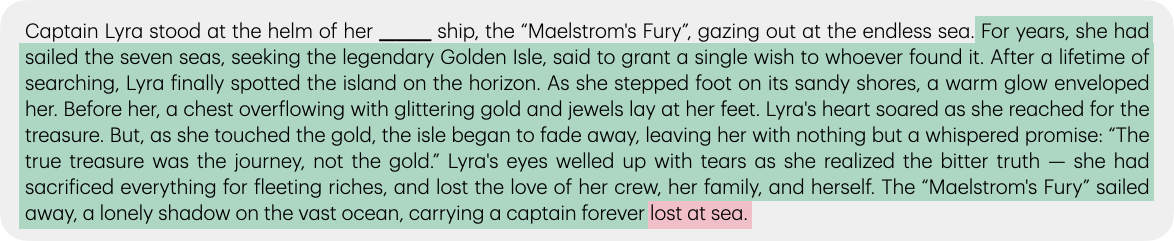}}
    \caption{\textbf{Comparison between the factual story and counterfactual variants.} Panel (a) shows the same factual story as in Section~\ref{sec:stories}. Panels (b, c, d) show the story resulting from various interventions. In each case, the first sentence (non-highlighted text) is provided as input to the LLM, with the word(s) in bold (or left empty) representing the intervention. The remainder of the output is regenerated using counterfactual token generation. Text highlighted in green indicates the tokens of the output that are identical in the factual story and its counterfactual counterpart.
    Red-highlighted text indicates the differences. In all panels, the temperature parameter is set to $\tau=0.9$.}
    \label{fig:stories-app}
\vspace{-5mm}
\end{figure}
\clearpage
\newpage


\section{Additional details on the experimental setup of Section~\ref{sec:bias} }\label{app:bias-setup}



In this section, we provide additional details about the census generation experiment discussed in Section~\ref{sec:bias}.

\begin{figure}[h]
\includegraphics[width=\linewidth]{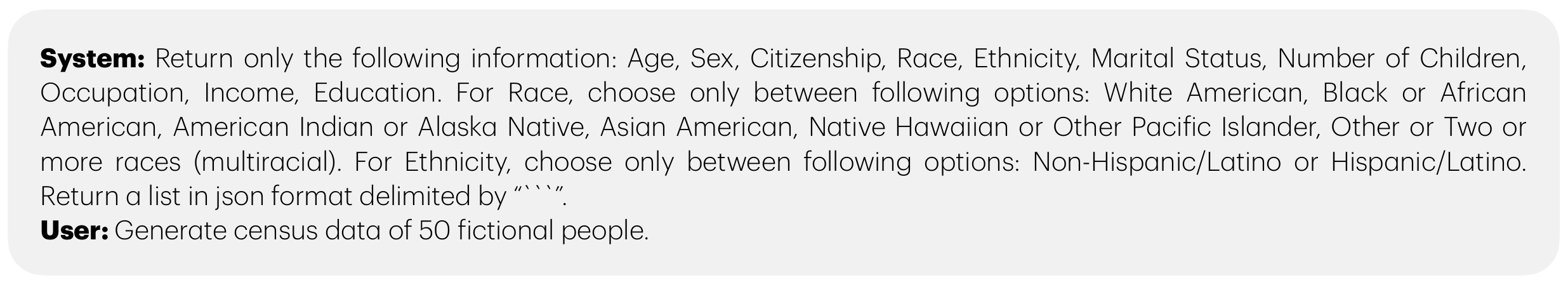}
\caption{The prompt used for census data generation.}~\label{fig:prompt}
\end{figure}

\begin{figure}[h]
    \centering
    \includegraphics[width=0.6\linewidth]{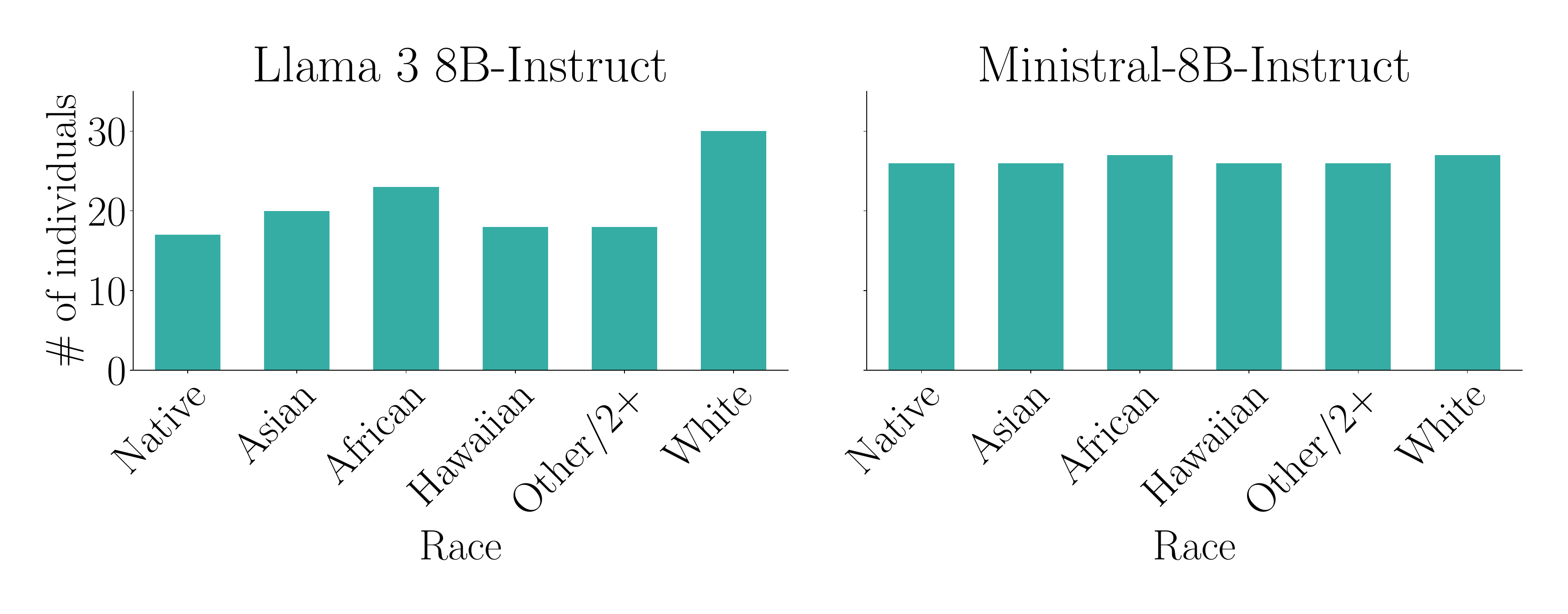}
    \caption{\textbf{Factual distributions of race.}
    The panels show the empirical factual distributions of race for the individuals generated by \texttt{Llama 3 8B-Instruct} (left) and \texttt{Ministral-8B-Instruct} (right).
    }
    \label{fig:race-factual}
\end{figure}

\xhdr{Data generation}
Figure~\ref{fig:prompt} shows the complete system and user prompts used to generate the census data. 
For race and ethnicity, we instructed the two LLMs, through the system prompt, to select values among those reported in the latest (2020) US Census.
We used this prompt three times with different seeds.
Despite our request for $50$ individuals per generation, \texttt{Llama 3 8B-Instruct} only generated $45$, $34$ and $48$ individuals each time, resulting in a total of $127$ individuals,
and \texttt{Ministral-8B-Instruct} generated $72$, $42$ and $44$ individuals each time, resulting in a total of $158$ individuals.
The census data for \texttt{Llama 3 8B-Instruct} included individuals for whom the generated income is exactly zero, which we discarded, resulting in a total of $114$ individuals. 
Additionally, for some interventions, the models generated incomplete counterfactual outputs; we excluded those cases from further analysis.
Table~\ref{tab:sex-factual} and Figure~\ref{fig:race-factual} show the factual distributions of sex and race for the individuals generated by each model.

\xhdr{Description of race values and education level}
Table~\ref{tab:races-description} contains the full description of each race attribute value, of which shortened versions were used in Figure~\ref{fig:race-education}.
Table~\ref{tab:education} lists the numerical values assigned to the (categorical) education attribute values, which were used to compute the difference in education level shown in Figure~\ref{fig:race-education}.

\begin{table}[h]
\centering
\begin{tabular}{@{}ll@{}}
\toprule
 Short & Full \\ \midrule
 Native &  American Indian or Alaska Native\\
 Asian &  Asian American\\
 African &  Black or African American\\
 Hawaiian &  Native Hawaiian or Other Pacific Islander\\
 Other/2+& Other or Two or more races (multiracial) \\ 
 White & White American \\
 \bottomrule
\end{tabular}
\caption{Short and full description of all races}\label{tab:races-description}
\end{table}

\begin{table}[h!]
\centering
\begin{tabular}{@{}ll@{}}
\toprule
 Education & Numerical Value \\ \midrule
 High school diploma, High school, & 1\\ High School Diploma, High School &  \\ \midrule
 Associate's degree, Associate's Degree, & \\ Associate degree, Associate Degree, & \\ Associate's, Associate, Undergraduate, & 2\\
 Some college, Some College, College,  &  \\
 Vocational Training  &  \\ \midrule
 Bachelor's degree, Bachelor's Degree, &  3\\
 Bachelor's, Nursing Degree &  \\ \midrule
 Master's degree, Master's Degree, &  4\\
 Master's &  \\ \midrule
 Ph.D., PhD, Doctorate degree, &   \\
 Doctorate Degree, Doctorate, &  \\
 Doctoral degree, Doctoral Degree, &  \\
 JD, Juris Doctor, Juris Doctor (JD), &  5\\
 Law degree, Law degree, PharmD, &  \\
 Pharmacy Degree, Dental degree, &  \\
 Dental Degree, Dentistry degree, &  \\
 MD, Medical degree, Medical Degree &  \\
 \bottomrule
\end{tabular}
\caption{Numerical value assigned to each (categorical) value of the education attribute}\label{tab:education}
\end{table}

\begin{table}[h]
\centering
\begin{tabular}{@{}lll@{}}
\toprule
 Model & Male & Female \\ \midrule
 \texttt{Llama 3 8B-Instruct} &  72 & 55\\
 \texttt{Ministral-8B-Instruct} &  79 & 79\\
 \bottomrule
\end{tabular}
\caption{Number of male and female individuals generated by each model}\label{tab:sex-factual}
\end{table}

\clearpage
\newpage

\section{Additional experimental results on bias detection} \label{app:bias-plots}


\begin{figure}[t]
    \captionsetup[subfigure]{justification=centering}
    \centering
    \includegraphics[width=0.3\textwidth]{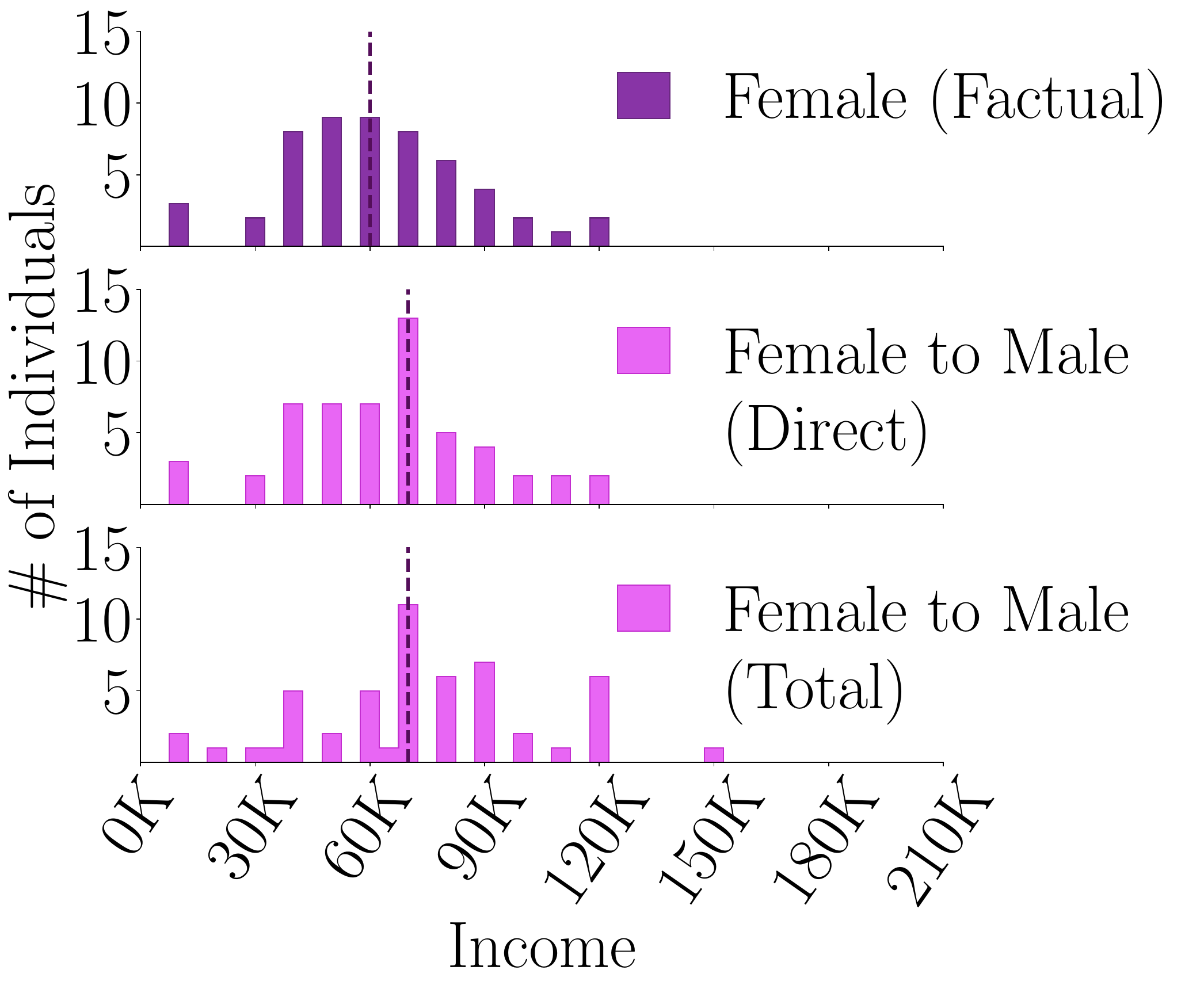}
    \hspace{2mm}
    \includegraphics[width=0.29\textwidth]{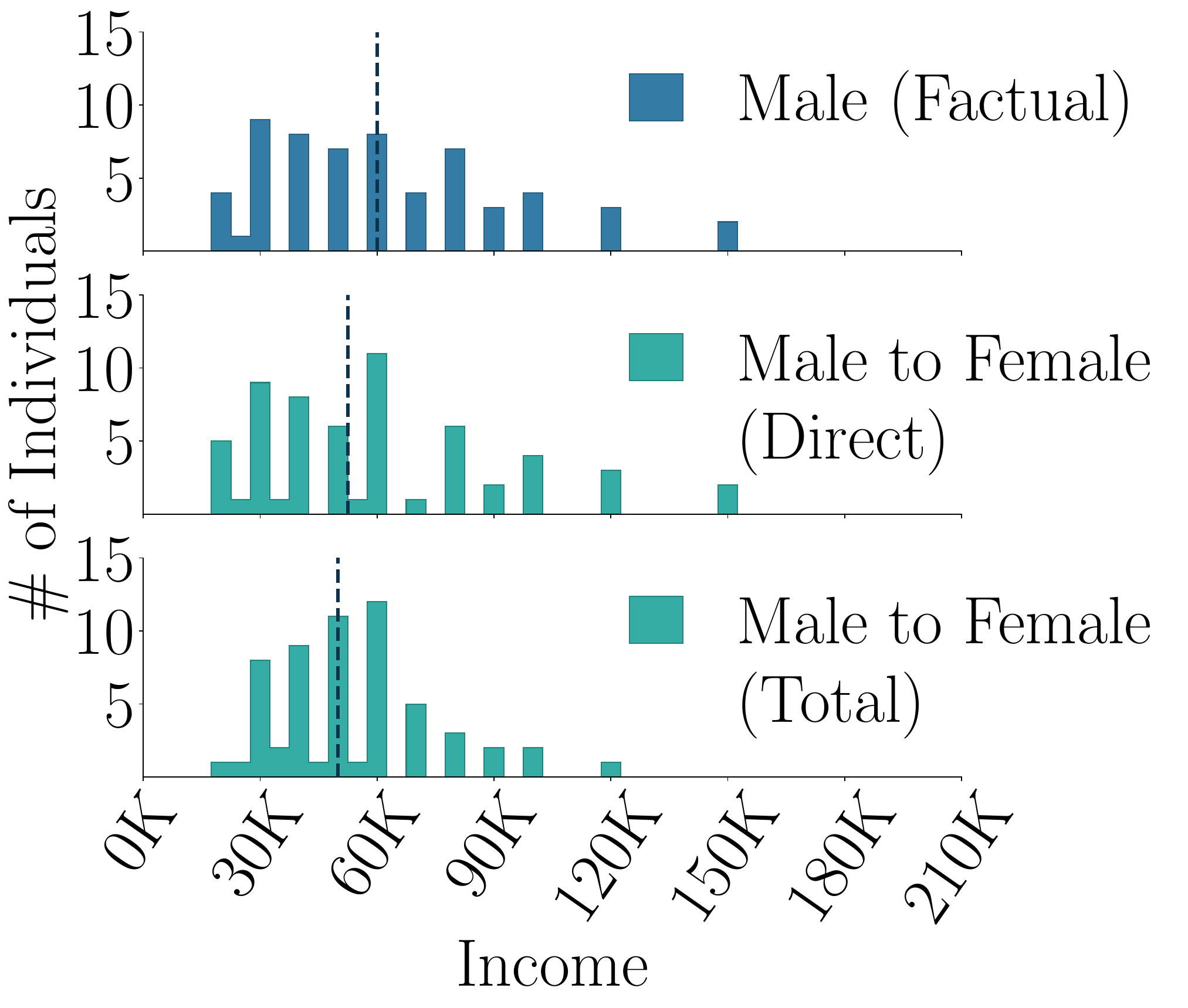}
        
    \caption{
    \textbf{Comparison between factual and counterfactual income.}
    %
    %
    The left (right) panel shows the factual distributions of income of female (male) individuals and their counterfactual 
    distribution of income under two interventions: (i) had they been male (female), while keeping fixed the rest of their attributes preceding income in the output sequence, and (ii) had they been male (female), while keeping fixed the attributes preceding sex but allowing the attributes between sex and income to change in the output sequence.
    In both panels, dashed lines correspond to the median income. Here, we  use \texttt{Llama 3 8B-Instruct} and set the temperature parameter to $\tau=0.8$.
    }\label{fig:bias-income-hist-llama}
    \vspace{-4mm}
\end{figure}

\begin{figure}[t]
    \captionsetup[subfigure]{justification=centering}
    \centering
    \subcaptionbox{Change in income upon intervention on sex (direct effect)~\label{fig:sex-income-direct-mistral}}{
        \includegraphics[width=0.3\textwidth]{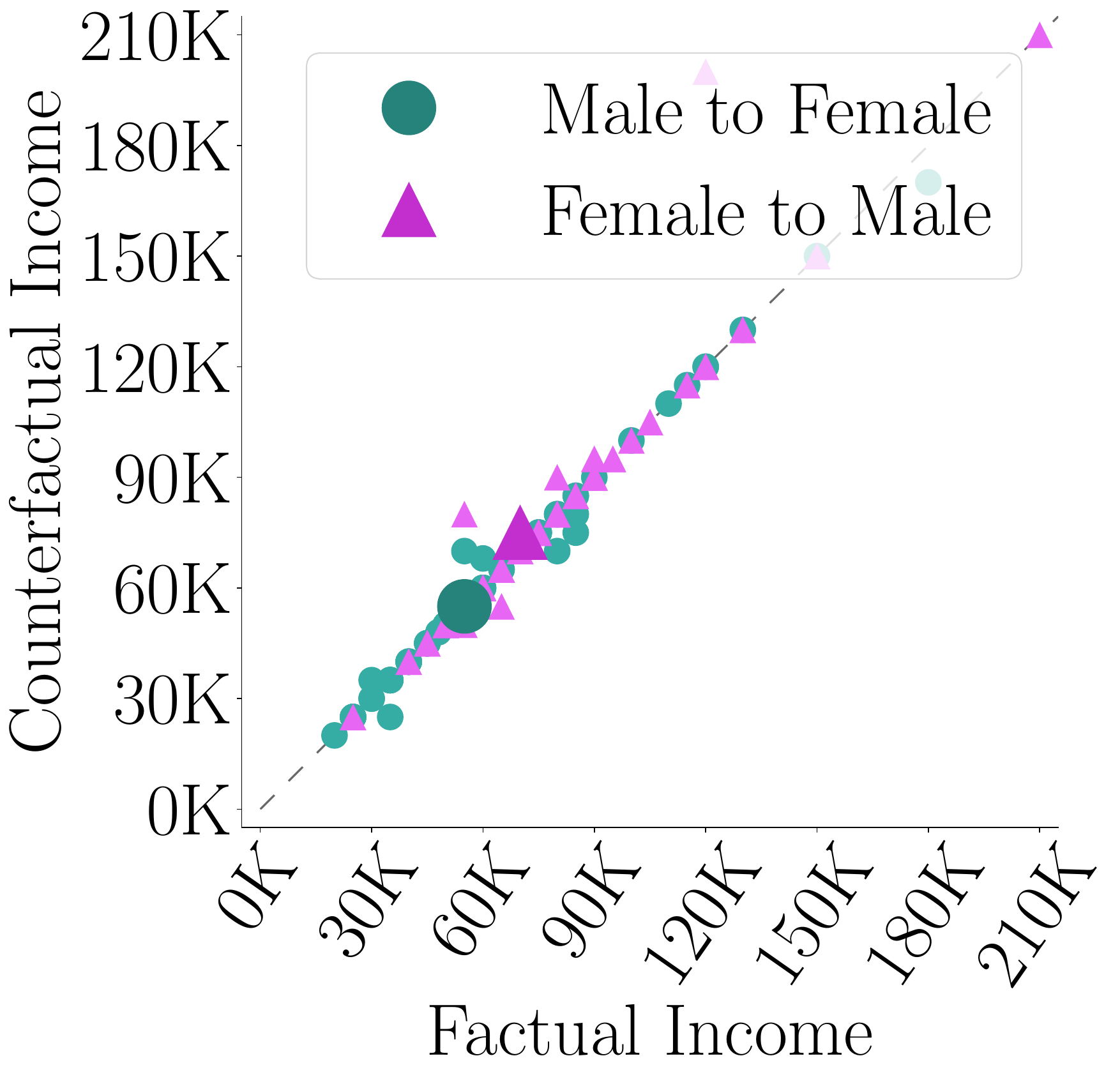}
    }
    \subcaptionbox{Change in income upon intervention on sex (total effect)~\label{fig:sex-income-total-mistral}}{
        \includegraphics[width=0.3\textwidth]{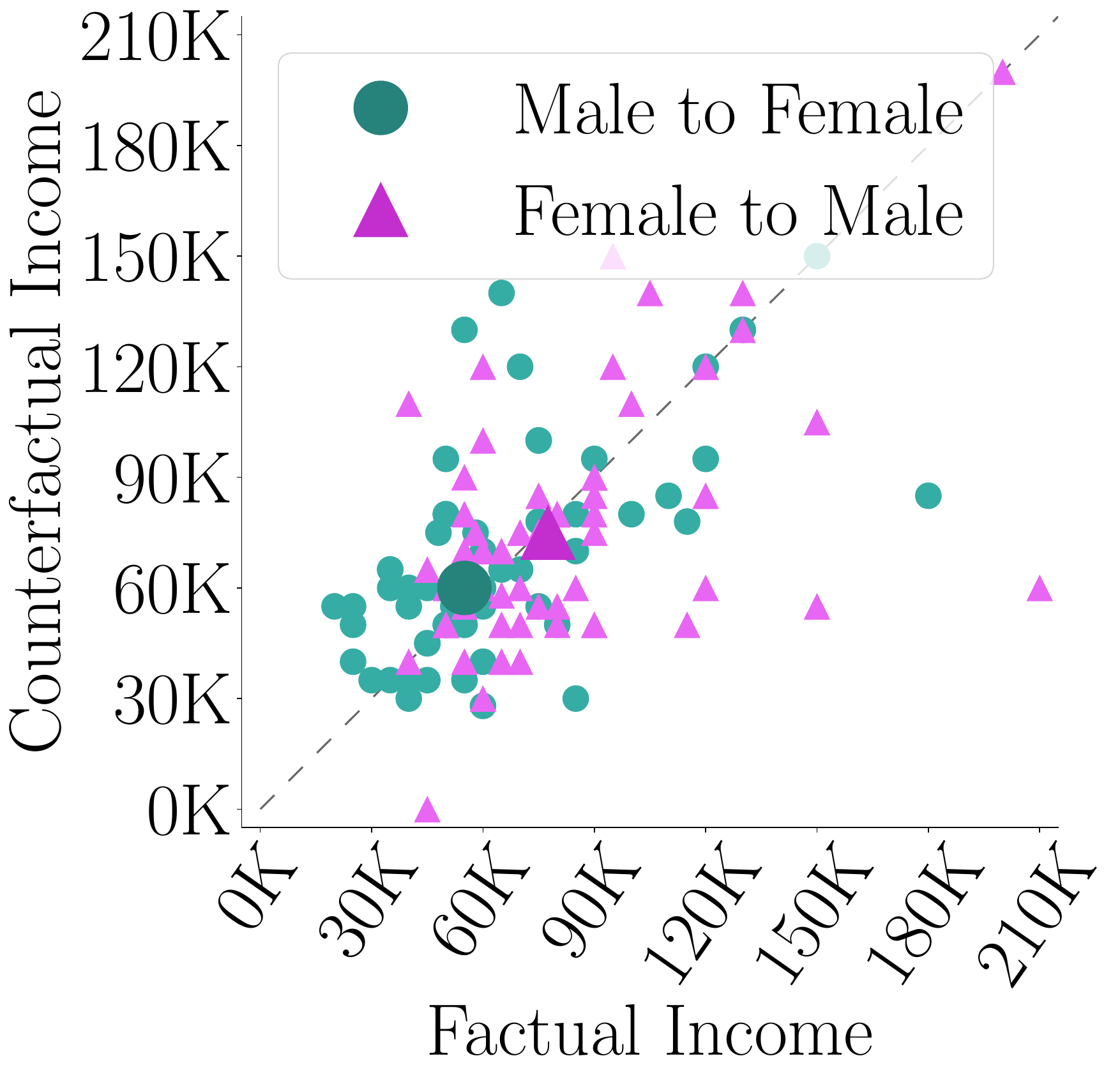}
    }
    \vspace{3mm}
    \\
    \subcaptionbox{Distributions of factual and counterfactual income~\label{fig:sex-income-hist-fem-mistral}}{
        \includegraphics[width=0.3\textwidth]{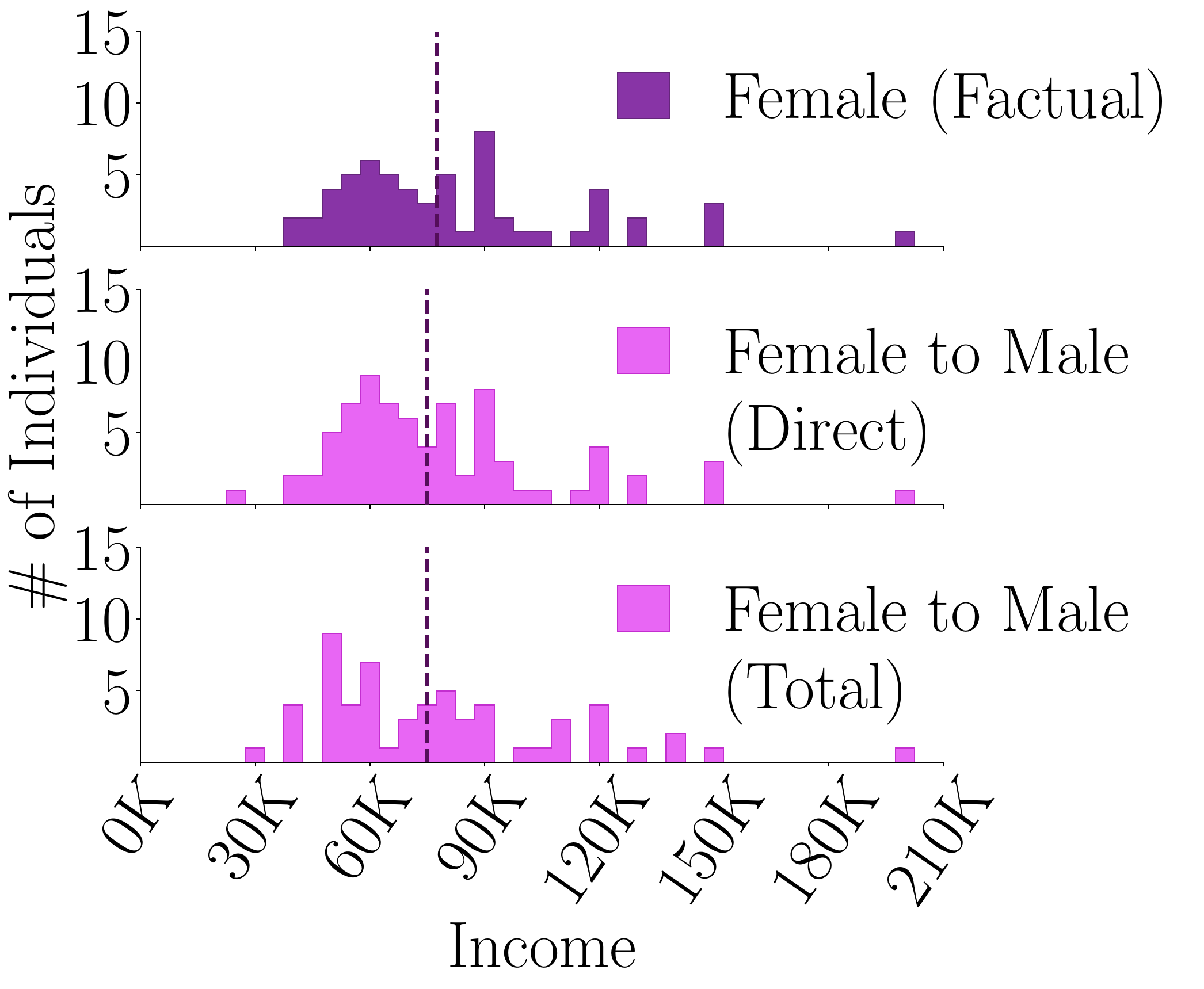}
    \hspace{2mm}
    \includegraphics[width=0.29\textwidth]{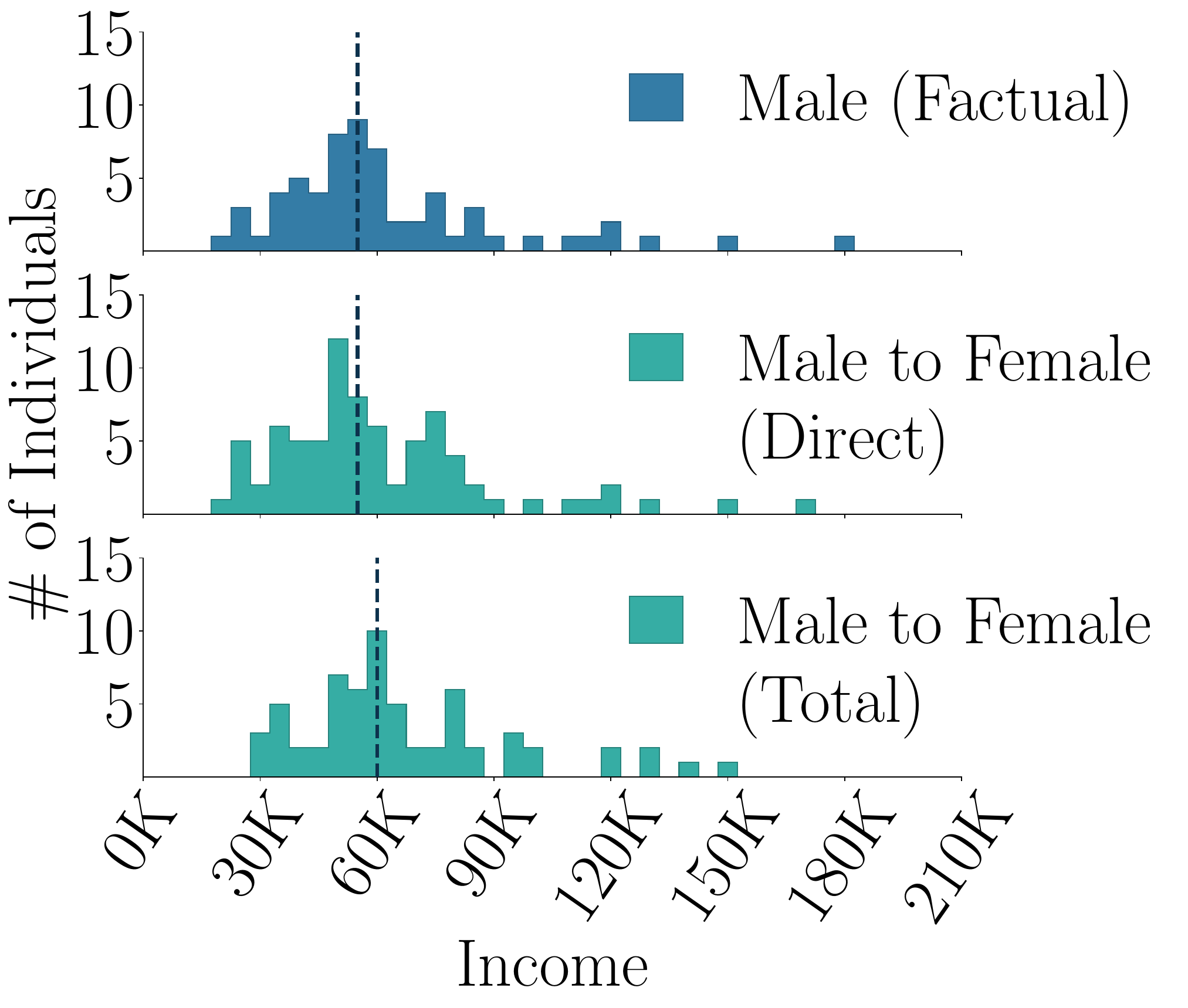}
    }
    \caption{
    \textbf{Comparison between factual and counterfactual income.}
    Panel (a) shows the change in income of male (female) individuals had they been female (male), while keeping fixed the rest of their attributes preceding income in the output sequence.
    Panel (b) shows the change in income of male (female) individuals had they been female (male), while keeping fixed the attributes preceding sex but allowing the attributes between sex and income to change in the output sequence.
    %
    Panel (c) shows the factual distributions of income of female and male individuals and their counterfactual distributions of income under the same interventions as in panels (a) and (b).
    Enlarged points in panels (a, b) and dashed lines in panel (c) correspond to the median income.
    In all panels, we use \texttt{Ministral-8B-Instruct} and set the temperature parameter to $\tau=0.8$.
    }\label{fig:bias-income-mistral}
    \vspace{-4mm}
\end{figure}

\begin{figure}[b]
    \captionsetup[subfigure]{justification=centering}
    \centering
    \subcaptionbox{Change in education level upon intervention on race~\label{fig:race-education-llama3}}{
        \includegraphics[width=0.33\textwidth]{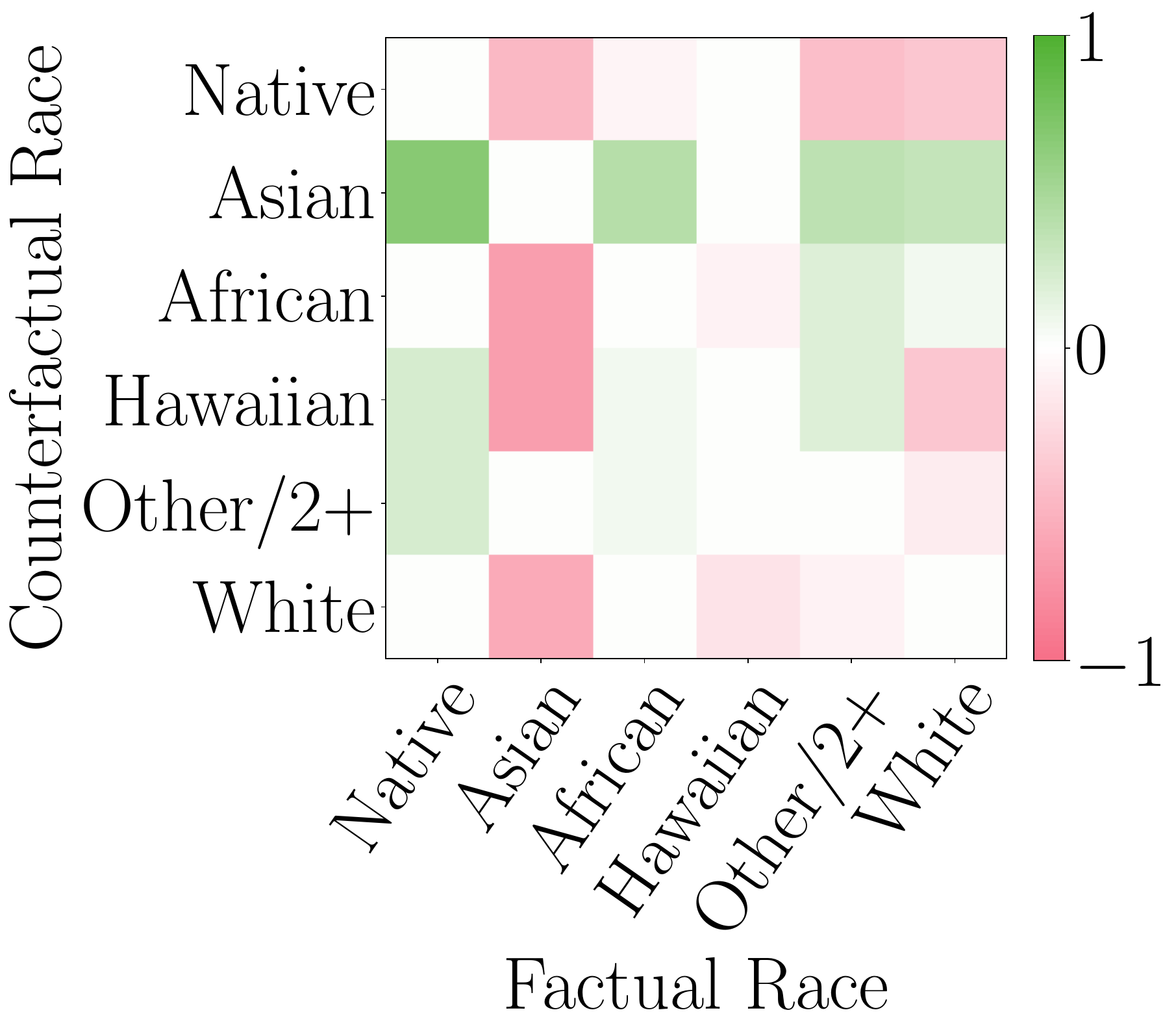}
    }
    \subcaptionbox{Change in occupation upon intervention on race~\label{fig:race_occupation-llama3}}{
        \includegraphics[width=0.5\textwidth]{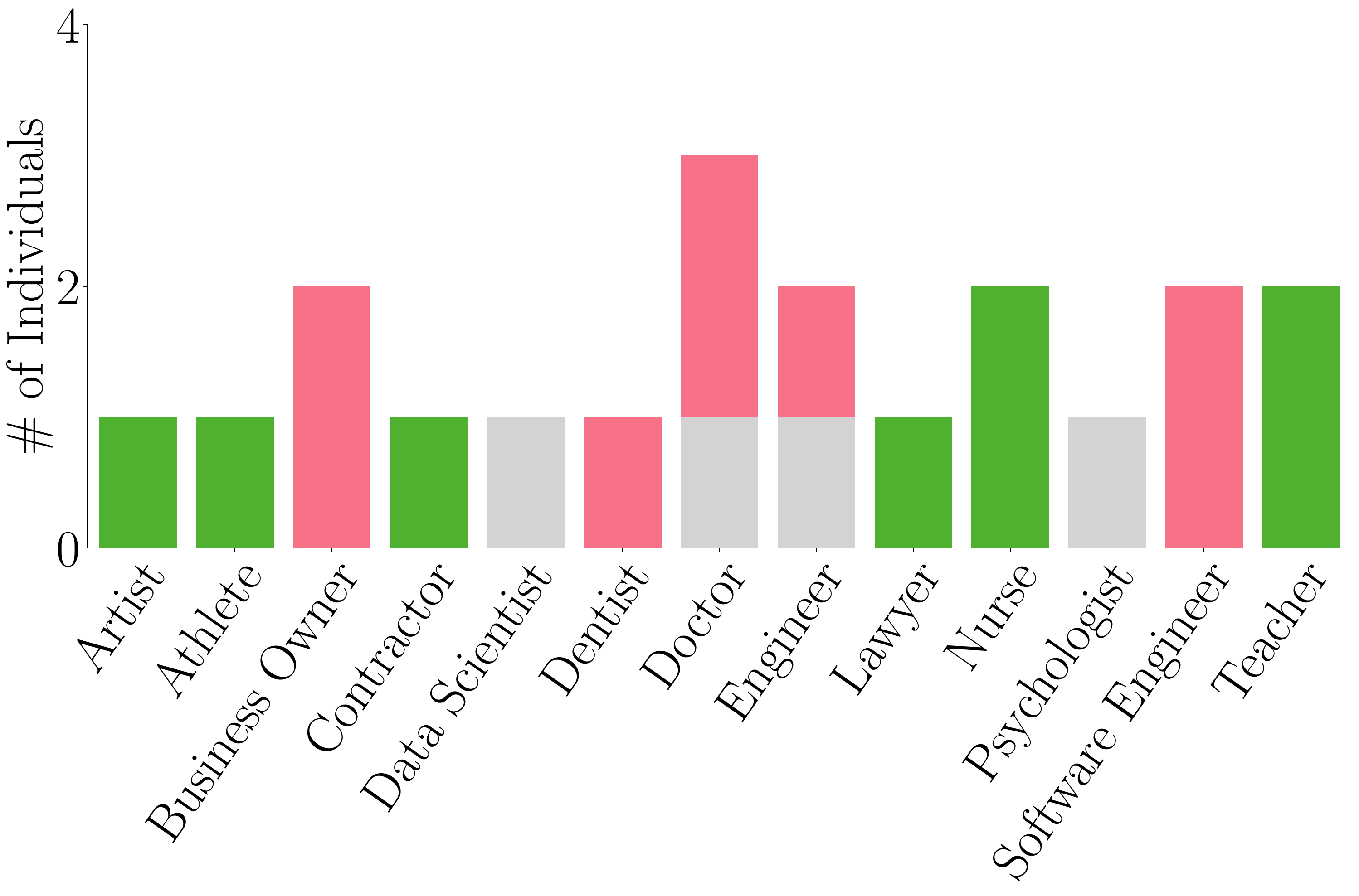}
    }
    
    \caption{
    \textbf{Comparison between factual and counterfactual education and occupation.}
    Panel (a) shows the average difference in the education level of individuals of each race had their race been different.
    Here, positive values indicate an improvement in education and negative values indicate a decline.
    Panel (b) shows the distribution shift of occupations among Asian American individuals had they been Black or African American. 
    Green (red) sections indicate the counterfactual increase (decrease) in the number of individuals that practice each occupation.
    In both panels, we use \texttt{Llama 3 8B-Instruct} and set the temperature parameter to $\tau=0.8$.}
    \label{fig:bias-education-occupation-llama3}
\end{figure}

Here, we present experimental results complementing those in Section~\ref{sec:bias}, regarding the direct and total effects of sex on income and the total effects of race on education and occupation. 
Figure~\ref{fig:bias-income-hist-llama} complements Figure~\ref{fig:sex-income-hist-fem}; it shows the distributions of the factual and counterfactual incomes of the individuals generated by \texttt{Llama 3 8B-Instruct}, under all possible interventions on their sex.
%
Figure~\ref{fig:bias-income-mistral} complements Figure~\ref{fig:bias-income}; it summarizes the results regarding direct and total effects of sex on income for the individuals generated by \texttt{Ministral-8B-Instruct}.
%
Figure~\ref{fig:bias-education-occupation-llama3} complements Figure~\ref{fig:bias-education-occupation}; it summarizes the results regarding total effects of race on education and occupation for the individuals generated by \texttt{Llama 3 8B-Instruct}.




\clearpage
\newpage

\section{Additional experiments using a sampler that does not satisfy counterfactual stability} \label{app:sampler}

In this section, we conduct counterfactual token generation 
using a classical sampler for categorical distributions, which can be viewed as an SCM that is not guaranteed to satisfy counterfactual stability.

\xhdr{Experimental setup}
We experiment with two different SCMs to implement the function $f_T$ responsible for sampling the next token on \texttt{Llama 3 8B-Instruct}: (i) the Gumbel-Max SCM defined in Eq.~\ref{eq:sampling-mechanism} and (ii) an SCM based on inverse transform sampling defined as follows:
\begin{equation}\label{eq:inverse-transform}
    f_T( D_i, U_i) = 
        \argmin_{j \in \{1, 2, \ldots, |V|\}} \left\{j\cdot \mathds{1}\left[\sum_{k = 1}^{j} D_{i,k} \geq U_i \right]\right\}.
\end{equation}
Under the SCM based on inverse transform sampling, the next token is sampled as follows. Let the tokens $t\in V$ follow a fixed order across time steps $i$. To generate a token $T_i$, for each position $j\in\{1,2,\ldots,|V|\}$, the SCM based on inverse transform sampling computes the cumulative sum of probabilities in the distribution $D_i$ corresponding to the tokens in positions $k\leq j$. Then, it samples a unidimensional noise variable $U_i\sim\text{Uniform}(0,1)$ and selects the first token in the vocabulary $V$ whose corresponding cumulative sum is greater than or equal to $U_i$. 

Similarly as in Section~\ref{sec:similarity}, we compare the edit distance between factual output sequences and sequences generated through interventional and counterfactual token generation using the Gumbel-Max SCM and the SCM given by Eq.~\ref{eq:inverse-transform}.

\begin{figure}[t]
    \centering
    \includegraphics[width=0.8\linewidth]{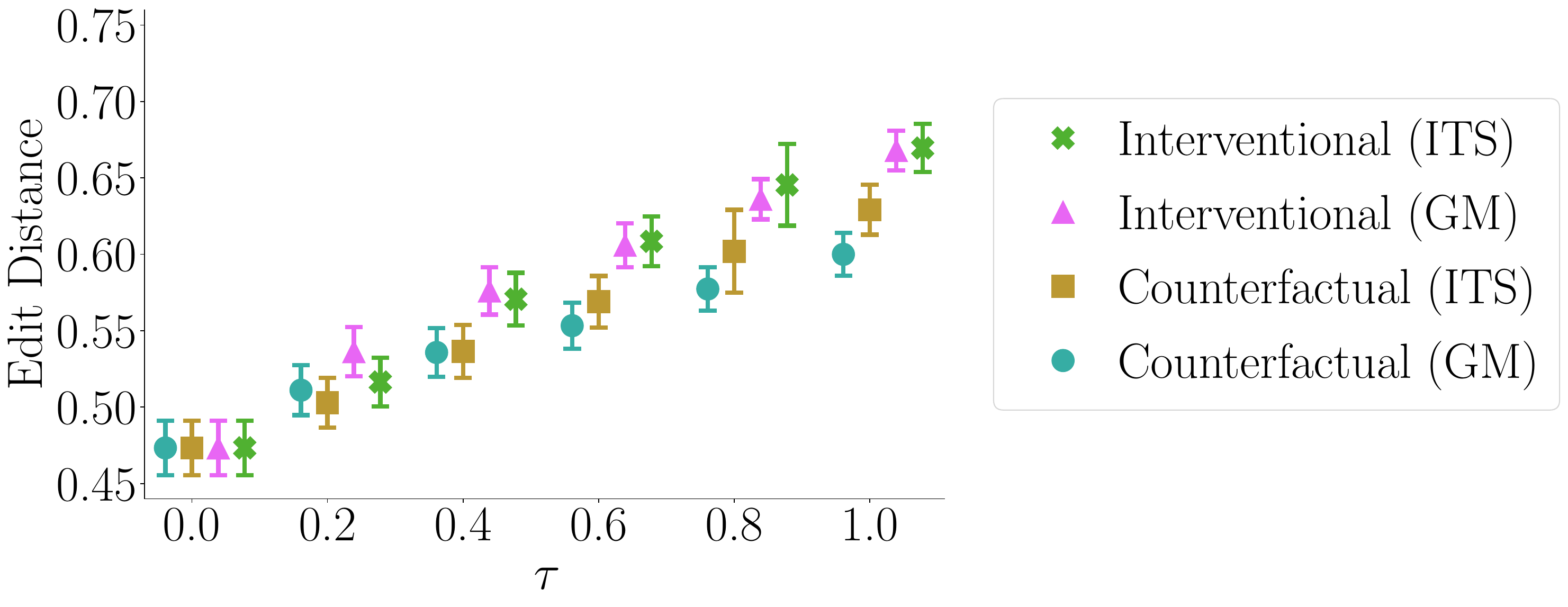}
    \caption{\textbf{Comparison between interventional and counterfactual token generation under two different SCMs.} The plot shows the edit distance between the factual sequence of tokens and the sequence generated by interventional and counterfactual token generation using the Gumbel-Max SCM (GM) defined by Eq.~\ref{eq:sampling-mechanism} and the SCM based on inverse transform sampling (ITS) defined by Eq.~\ref{eq:inverse-transform}, against various values of the temperature parameter $\tau$.
    The edit distance is averaged over $4{,}000$ output sequences, resulting from two independent interventions per factual sequence, and error bars represent $95\%$ confidence intervals.
    In this panel, we use \texttt{Llama 3 8B-Instruct}.}
    \label{fig:non-gumbel-scm}
\end{figure}

\xhdr{Results} Figure~\ref{fig:non-gumbel-scm} summarizes the results, which show that, under both SCMs, the output sequences generated using counterfactual token generation remain more similar to their respective factual sequence (\ie,  exhibit lower edit distance) compared to the sequences generated using interventional token generation. 
Moreover, as expected, the counterfactual output sequences generated using the SCM defined by Eq.~\ref{eq:inverse-transform}, which does not satisfy the property of counterfactual stability, exhibit an edit distance higher than those generated using the Gumbel-Max SCM.

